\RequirePackage{fix-cm}
\documentclass[smallextended]{svjour3}       
\smartqed  
\usepackage{amsmath}
\usepackage{amssymb}
\usepackage{booktabs}
\usepackage{comment}
\usepackage{graphicx}
\usepackage{todonotes}
\usepackage{hyperref}
\usepackage{multirow}
\usepackage[numbers]{natbib}
\usepackage{subfig}
\usepackage{collcell}
\usepackage{tikz}
\usetikzlibrary{arrows,arrows.meta,patterns,automata,backgrounds,decorations,fit,petri,positioning,petri,shapes,calc,shadows}

\newcommand{\fleche}{\longrightarrow}
\newcommand{\flsup}[1]{\stackrel{#1}{\fleche}}
\newcommand{\step}[1]{\flsup{#1}}           
\newcommand{\Lan}                 {\mathfrak{L}}

\newcommand{\APN}{\mathit{APN}}

\newlength{\hatchspread}
\newlength{\hatchthickness}
\newlength{\hatchshift}
\newcommand{\hatchcolor}{}
\tikzset{hatchspread/.code={\setlength{\hatchspread}{#1}},
	hatchthickness/.code={\setlength{\hatchthickness}{#1}},
	hatchshift/.code={\setlength{\hatchshift}{#1}},
	hatchcolor/.code={\renewcommand{\hatchcolor}{#1}}}
\tikzset{hatchspread=3pt,
	hatchthickness=0.4pt,
	hatchshift=0pt,
	hatchcolor=black}
\pgfdeclarepatternformonly[\hatchspread,\hatchthickness,\hatchshift,\hatchcolor]
{custom north west lines}
{\pgfqpoint{\dimexpr-2\hatchthickness}{\dimexpr-2\hatchthickness}}
{\pgfqpoint{\dimexpr\hatchspread+2\hatchthickness}{\dimexpr\hatchspread+2\hatchthickness}}
{\pgfqpoint{\dimexpr\hatchspread}{\dimexpr\hatchspread}}
{
	\pgfsetlinewidth{\hatchthickness}
	\pgfpathmoveto{\pgfqpoint{0pt}{\dimexpr\hatchspread+\hatchshift}}
	\pgfpathlineto{\pgfqpoint{\dimexpr\hatchspread+0.15pt+\hatchshift}{-0.15pt}}
	\ifdim \hatchshift > 0pt
	\pgfpathmoveto{\pgfqpoint{0pt}{\hatchshift}}
	\pgfpathlineto{\pgfqpoint{\dimexpr0.15pt+\hatchshift}{-0.15pt}}
	\fi
	\pgfsetstrokecolor{\hatchcolor}
	\pgfusepath{stroke}
}

\newcommand*{\MinNumberSepsis}{0.0}%
\newcommand*{\MaxNumberSepsis}{0.0846}%
\newcommand{\ApplyGradientSepsis}[1]{%
    \pgfmathsetmacro{\PercentColor}{(100*(1-(#1-\MinNumberSepsis)/(\MaxNumberSepsis-\MinNumberSepsis)))}
    \hspace{-0.33em}\colorbox{green!\PercentColor!red}{#1}
}
\newcolumntype{s}{>{\collectcell\ApplyGradientSepsis}{r}<{\endcollectcell}}

\newcommand*{\MinNumberSepsiss}{0.0}%
\newcommand*{\MaxNumberSepsiss}{0.0103}%
\newcommand{\ApplyGradientSepsiss}[1]{%
	\pgfmathsetmacro{\PercentColor}{(100*(1-(#1-\MinNumberSepsiss)/(\MaxNumberSepsiss-\MinNumberSepsiss)))}
	\hspace{-0.33em}\colorbox{green!\PercentColor!red}{#1}
}
\newcolumntype{S}{>{\collectcell\ApplyGradientSepsiss}{r}<{\endcollectcell}}

\newcommand*{\MinNumberBpi}{0.0}%
\newcommand*{\MaxNumberBpi}{0.0505}%
\newcommand{\ApplyGradientBpi}[1]{%
    \pgfmathsetmacro{\PercentColor}{(100.0*(1-(#1-\MinNumberBpi)/(\MaxNumberBpi-\MinNumberBpi)))}
    \hspace{-0.33em}\colorbox{green!\PercentColor!red}{#1}
}
\newcolumntype{b}{>{\collectcell\ApplyGradientBpi}{r}<{\endcollectcell}}

\newcommand*{\MinNumberBpis}{0.0}%
\newcommand*{\MaxNumberBpis}{0.006}%
\newcommand{\ApplyGradientBpis}[1]{%
	\pgfmathsetmacro{\PercentColor}{(100.0*(1-(#1-\MinNumberBpis)/(\MaxNumberBpis-\MinNumberBpis)))}
	\hspace{-0.33em}\colorbox{green!\PercentColor!red}{#1}
}
\newcolumntype{B}{>{\collectcell\ApplyGradientBpis}{r}<{\endcollectcell}}

\newcommand*{\MinNumberReceipt}{0.0}%
\newcommand*{\MaxNumberReceipt}{0.0392}%
\newcommand{\ApplyGradientReceipt}[1]{%
    \pgfmathsetmacro{\PercentColor}{(100.0*(1-(#1-\MinNumberReceipt)/(\MaxNumberReceipt-\MinNumberReceipt)))}
    \hspace{-0.33em}\colorbox{green!\PercentColor!red}{#1}
}
\newcolumntype{w}{>{\collectcell\ApplyGradientReceipt}{r}<{\endcollectcell}}

\newcommand*{\MinNumberReceipts}{0.0}%
\newcommand*{\MaxNumberReceipts}{0.0042}%
\newcommand{\ApplyGradientReceipts}[1]{%
	\pgfmathsetmacro{\PercentColor}{(100.0*(1-(#1-\MinNumberReceipts)/(\MaxNumberReceipts-\MinNumberReceipts)))}
	\hspace{-0.33em}\colorbox{green!\PercentColor!red}{#1}
}
\newcolumntype{W}{>{\collectcell\ApplyGradientReceipts}{r}<{\endcollectcell}}

\newcommand*{\MinNumberNasa}{0.0}%
\newcommand*{\MaxNumberNasa}{0.0233}%
\newcommand{\ApplyGradientNasa}[1]{%
	\pgfmathsetmacro{\PercentColor}{(100.0*(1-(#1-\MinNumberNasa)/(\MaxNumberNasa-\MinNumberNasa)))}
	\hspace{-0.33em}\colorbox{green!\PercentColor!red}{#1}
}
\newcolumntype{n}{>{\collectcell\ApplyGradientNasa}{r}<{\endcollectcell}}

\newcommand*{\MinNumberNasas}{0.0}%
\newcommand*{\MaxNumberNasas}{0.0044}%
\newcommand{\ApplyGradientNasas}[1]{%
	\pgfmathsetmacro{\PercentColor}{(100.0*(1-(#1-\MinNumberNasas)/(\MaxNumberNasas-\MinNumberNasas)))}
	\hspace{-0.33em}\colorbox{green!\PercentColor!red}{#1}
}
\newcolumntype{N}{>{\collectcell\ApplyGradientNasas}{r}<{\endcollectcell}}

\begin{document}
\sloppy

\title{An Interdisciplinary Comparison of Sequence Modeling Methods for Next-Element Prediction 
}

\titlerunning{Benchmarking Sequence Modeling Methods for Next-Element Prediction}        

\author{Niek Tax \and
		Irene Teinemaa \and
		Sebastiaan J. van Zelst         
}


\institute{
           Niek Tax \at
           Department of Mathematics and Computer Science, Eindhoven University of Technology,\\ 
           P.O. Box 513, 5600 MB, Eindhoven, The Netherlands\\
           \email{n.tax@tue.nl}
           \and
          Irene Teinemaa \at
           Institute of Computer Science, University of Tartu,\\
           J Liivi 2, 50409 Tartu, Estonia\\
           \email{irene.teinemaa@ut.ee}
           \and
           Sebastiaan J. van Zelst \at
           Fraunhofer Institute for Applied Information Technology, FIT,\\ 
           Konrad-Adenauer-Straße, 53754 Sankt Augustin, Germany\\
           \email{sebastiaan.van.zelst@fit.fraunhofer.de}           
}

\date{Received: date / Accepted: date}

\maketitle

\begin{abstract}
Data of sequential nature arise in many application domains in forms of, e.g. textual data, DNA sequences, and software execution traces. 
Different research disciplines have developed methods to learn \emph{sequence models} from such datasets: (i) in the \emph{machine learning} field methods such as (hidden) Markov models and recurrent neural networks have been developed and successfully applied to a wide-range of tasks, (ii) in \emph{process mining} process discovery techniques aim to generate human-interpretable descriptive models, and (iii) in the \emph{grammar inference} field the focus is on finding descriptive models in the form of formal grammars.
Despite their different focuses, these fields share a common goal - learning a model that accurately describes the behavior in the underlying data. Those sequence models are \emph{generative}, i.e, they can predict what elements are likely to occur after a given unfinished sequence. So far, these fields have developed mainly in isolation from each other and no comparison exists. This paper presents an interdisciplinary experimental evaluation that compares sequence modeling techniques on the task of \emph{next-element prediction} on four real-life sequence datasets. 
The results indicate that machine learning techniques that generally have no aim at interpretability in terms of accuracy outperform techniques from the process mining and grammar inference fields that aim to yield interpretable models.
\end{abstract}

\section{Introduction}
A large share of the world's data naturally occurs in sequences. 
Examples thereof include textual data (e.g., sequences of letters or of words), DNA sequences, web browsing behavior, and execution traces of business processes or of software systems. 
Several different research fields have focused on the development of tools and techniques to model and describe such \emph{sequence data}.
However, these research fields mostly operate independently from each other, and, with little knowledge transfer between them.
Nonetheless, the different techniques from the different research fields generally share the same common goal: learning a descriptive model from a dataset of sequences such that the model accurately \emph{generalizes} from the sequences that are present in the dataset. 
The three major research communities that have developed sequence modeling techniques include the fields of \emph{machine learning}, \emph{grammar inference}, and \emph{process mining}.

In the \emph{machine learning} research community several techniques for modeling sequence have been developed in the \emph{sequence modeling} subfield. 
Well-known examples of such sequence models include n-gram models~\cite{Dunning1994}, Markov chains~\cite{gagniuc2017markov}, Hidden Markov Models (HMMs)~\cite{Rabiner1989} and Recurrent Neural Networks (RNNs)~\cite{Hopfield1982}. 
Sequence modeling techniques have been successfully applied in many application domains, including modeling of natural language~\cite{Dunning1994}, music sequences~\cite{Logan2000}, and DNA sequences~\cite{Stanke2003}. 
Sequence modeling techniques from the machine learning field  typically have as end-goal to automate a certain task and therefore focus mostly on the accuracy of the models and put little emphasis on producing human-interpretable models. Examples of successfully automated tasks within aforementioned application domains include automated translation of text documents~\cite{Cho2014}, and automatic music generation and music transcription~\cite{Boulanger2012}.

In the field of \emph{grammar inference}~\cite{Higuera2005,Higuera2010} it is typically assumed that there exists an unknown formal language that generates the sequential data set.
As such, the sequences in the dataset are considered \emph{positive examples}, on the basis of which the resulting formal language is to be inferred. 
The language that is learned is represented as an automaton in the case the language is regular or alternatively as a context-free grammar, thereby contrasting the machine learning field by creating human-interpretable models.

The \emph{process mining}~\cite{Aalst2016} field, originating from the research domain of \emph{business process management}, is concerned with finding an accurate description of a business process from a dataset of logged execution sequences, captured during the execution of the process.
The result of such process mining algorithm is usually a process model, i.e., a model in a graphical form with a corresponding formal semantics. 
Some of the process model notations that are generated by process mining techniques are heavily used in industry, e.g. BPMN~\cite{bpmn_2011}. 
As such, in the process mining field, there is a strong focus on the human-interpretability of the discovered models. 
Another feature that differentiates process mining from the machine learning and grammar inference techniques is its focus on modeling of concurrent behavior explicitly. 
Concurrent execution of tasks plays such an important role in business processes that process mining techniques aim to model concurrency explicitly.

There have been efforts to systematically compare and benchmark the accuracy of within each of these research fields individually. In the machine learning field this often occurs through task-specific benchmark datasets, such as the popular WMT'14 dataset for machine translation~\cite{Bojar2014}. In the grammar inference this often happens through competitions with standardized evaluation~\cite{Balle2017,Clark2007,Lang1998}. In the process mining field~\cite{augusto2018automated} compares 35 process discovery algorithms on a collection of datasets, and additionally competitions with standardized evaluation has emerged in the process mining field~\cite{Carmona2016}. 

While there have been many efforts to compare sequence modeling techniques \emph{within} each research field, to the best of our knowledge, there has been little to no work in the evaluating the accuracy of sequence models \emph{between} the different research fields. 
In order to enable a comparison between sequence models from the different research fields we focus on a single task to which such models can be applied: predicting the next element of an unfinished sequence. More specifically, we focus on generating the whole probability distribution over possible next elements of the sequence, instead of just predicting the single most likely next element. Machine learning approaches are generally already capable of generating the whole probability distribution over possible next elements and the same holds for a subset of grammar inference techniques, called \emph{probabilistic grammar inference}. In earlier work~\cite{DBLP:conf/caise/TaxZT18}, we presented a method to use a process model as a \emph{probabilistic sequence classifier}, thereby enabling the comparison with other probabilistic sequence models. Furthermore,~\cite{DBLP:conf/caise/TaxZT18} presented a preliminary comparison between machine learning and process discovery techniques.

This paper extends the work started in~\cite{DBLP:conf/caise/TaxZT18} in several ways. 
First, we have expanded the scope of the paper to include a new research community that was not yet represented in the initial study: the field of \emph{grammar inference}. 
Secondly, we expanded our experimental setup by experimenting on a larger number of datasets and covering a larger set of techniques, additionally covering \emph{hidden Markov models}~\cite{Rabiner1989} and Active LeZi~\cite{Gopalratnam2007} in the machine learning category and in the process mining category adding the Indulpet Miner~\cite{Leemans2018} process discovery algorithm as well as a class of automaton-based prediction techniques. 
Finally, in an attempt to bring three research communities together and make this manuscript useful for researchers from the machine learning, process mining, as well as grammar inference domains, we have added considerable detail to the description of process-model-based predictions.

The remainder of this paper is structured as follows. 
In \autoref{sec:background}, we describe basic concepts and notations that are used throughout the paper. 
In \autoref{sec:prediction}, we present the different next element prediction techniques studied in this paper.
We describe the experimental setup in \autoref{sec:experimental_setup} and discuss the results of the experiments in \autoref{sec:results}. In \autoref{sec:related_work}, we present an overview of related work.
Finally, we conclude this paper and identify several interesting areas of future work in \autoref{sec:conclusion}.

\section{Preliminaries}
\label{sec:background}
In this section, we introduce preliminary concepts used in later sections of this paper.
We cover the fundamental basis of sequential data and, formalize the notion of process discovery and present the semantics of  process-oriented models that we use in this paper for the purpose of next-element prediction.

\subsection{Sequences and Multisets}
Sequences relate positions to elements, i.e. they allow us to order elements.
A sequence of length $n$ over a set $X$ is a function $\sigma \colon \{1,...,n\} \to X$ and is alternatively written as $\sigma{=}\langle \sigma(1),\sigma(2),\dots,\sigma(n)\rangle$.
$X^*$ denotes the set of all possible finite sequences over a set $X$.
Given a sequence $\sigma \in X^*$, $|\sigma|$ denotes its length, e.g. $|\langle x, y, z \rangle| = 3$.
We represent the empty sequence by $\epsilon$, i.e. $|\epsilon| = 0$, and, $\sigma_1{\cdot}\sigma_2$ denotes the concatenation of sequences $\sigma_1$ and $\sigma_2$. 
$\mathit{hd}^k(\sigma){=}\langle \sigma(1), \sigma(2), \dots, \sigma(k) \rangle$ is the prefix (or head) of length $k$ of sequence $\sigma$ (with $0 {<} k {<} |\sigma|$), e.g. $\mathit{hd}^2(\langle a,b,c,d,e\rangle){=}\langle a,b\rangle$. 
Similarly, $\mathit{tl}^k(\sigma){=} \langle \sigma(|\sigma| - k + 1), ..., \sigma(|\sigma|- 1), \sigma(|\sigma|) \rangle$ is the postfix (or tail) of length $k$ (with $0 {<} k {<} |\sigma|$).
Given $\sigma \in X^*$ and $1 \leq i \leq j \leq |\sigma|$, we let $\sigma_{(i,j)} = \langle \sigma(i), ..., \sigma(j) \rangle$. We say that one sequence $\sigma$ is a prefix of another sequence $\sigma'$ if and only if $\mathit{hd}^{|\sigma|}(\sigma')=\sigma$.

Given a function $f{\colon}X{\to}Y$ and a sequence $\sigma=\langle\sigma(1),...,\sigma(n)\rangle \in X^*$, we lift function application to sequences, i.e. $f(\sigma) \in Y^*$, where $f(\sigma) = \langle f(\sigma(1)), ..., f(\sigma(n)) \rangle$.
Furthermore, given $\sigma \in X^*$ and $X' \subseteq X$, we define $\sigma_{\downarrow_{X'}} \in X'^*$, with (1) $\epsilon_{\downarrow_{X'}} = \epsilon$ and (2) for any $\sigma \in X^*$ and $x \in X$:
\begin{center}
	$(\sigma \cdot \langle x\rangle)_{\downarrow_{X'}} =
	\left\{
	\begin{array}{ll}
	\sigma \cdot \langle x \rangle & \mbox{if\ } x \in X',\\
	\sigma  & \mbox{otherwise}. \\	
	\end{array}
	\right.$
\end{center}


A multiset (or bag) over $X$ is a function $B:X{\rightarrow}\mathbb{N}$ which we write as $[x_1^{w_1},x_2^{w_2},\dots,x_n^{w_n}]$, where for $1{\le} i {\le} n$ we have $x_i{\in} X$ and $w_i{\in}\mathbb{N}^{+}$.
Here, $w_i$ represents the value of $B$ for $x_i$, i.e. $B(x_i) = w_i$.
In case $w_i = 0$, we omit $x_i$ from multiset notation, and, in case $w_i = 1$, we simply write $x_i$, i.e. without $w_i$ as its superscript.
For example, for multiset $B_1 = [a,c^2]$ over set $X=\{a,b,c\}$, we have $B_1(a) = 1$, $B_1(b) = 0$ and $B_1(c) = 2$.
The empty multiset is written as $[\ ]$.
The set of all multisets over $X$ is denoted $\mathcal{B}(X)$.
Given $B \in \mathcal{B}(X)$, we let $\tilde{B} = \left\{ x \in X \mid B(x) > 0  \right\}$.

Finally, given $\sigma \in X^*$, we let $\overline{\sigma} = \left\{ x \in X \mid \exists{{i}{\in}{\{1,...,|\sigma|\}}}{\left(\sigma(i){=}x \right)}\right\}$ and $\overrightarrow{\sigma} = \left[{x^k \in X \mid k = |\{{i}{\in}{\{1,...,|\sigma|\}} \mid \sigma(i){=}x \} |}\right]$ (also known as the Parikh representation of the sequence), e.g. $\overline{\langle a,b,b,c \rangle} = \{a,b,c\}$ and $\overrightarrow{\langle a,b,b,c \rangle} = [a,b^2,c]$.

\subsection{Sequence Databases}

As indicated, in this paper, we study next-element prediction on the basis of \emph{sequential data}.
As an example introduction to this type of data, we present fictional data, which is assumed to be generated and captured during the execution of a \emph{(business) process}.
Consider \autoref{tab:example_log}, adopted from~\cite{Aalst2016}, in which we depict example data related to the process of handling a compensation request for concert tickets.
\begin{table}[tb]
	\caption{A fictional example sequence database, adopted from~\cite{Aalst2016}, describing behavior related to a compensation request process for concert tickets.}
	\label{tab:example_log}
	\begin{center}
		\resizebox{\linewidth}{!}{
		\begin{tabular}{lllllll}
			\toprule
			\emph{Case id} & \emph{Event id} & \emph{Timestamp} & \emph{Activity} & \emph{Resource} & \emph{Cost} & $\cdots$ \\
			\midrule
			$\vdots$ & $\vdots$ & $\vdots$ & $\vdots$ & $\vdots$ & $\vdots$ & $\cdots$\\
			$1337$ & $745632$ & 30-7-2018 11.02 & register request (a) & Barbara & 50 & $\cdots$\\
			$1338$ & $745633$ & 30-7-2018 11.32 & register request (a) & Jan & 50 & $\cdots$\\
			$1337$ & $745634$ & 30-7-2018 12.12 & check ticket (d) & Stefanie & 100 & $\cdots$\\
			$1338$ & $745635$ & 30-7-2018 14.16 & examine casually (c) & Jorge & 400 & $\cdots$\\
			$1339$ & $745636$ & 30-7-2018 14.32 & register request (a) & Josep & 50 & $\cdots$\\
			$1339$ & $745637$ & 30-7-2018 15.42 & examine thoroughly (b) & Marlon & 600 & $\cdots$\\
			$1337$ & $745638$ & 3-8-2018 11.18  & examine thoroughly (b) & Barbara & 600 & $\cdots$\\
			$1337$ & $745639$ & 3-8-2018 15.34  & decide (e) & Wil & 700 & $\cdots$\\
			$1338$ & $745640$ & 3-8-2018 15.50  & check ticket & Marcello & 100 & $\cdots$\\
			$1337$ & $745641$ & 3-8-2018 16.50  & reject request (h) & Arthur & 25 & $\cdots$\\
			$1340$ & $745642$ & 3-8-2018 16.58  & register request (a) & Hajo & 50 & $\cdots$\\
			$1338$ & $745643$ & 3-8-2018 16.59  & pay compensation (g) & Boudewijn & 75 & $\cdots$\\
			$\vdots$ & $\vdots$ & $\vdots$ & $\vdots$ & $\vdots$ & $\vdots$ & $\cdots$\\
			\bottomrule
		\end{tabular}}
	\end{center}
\end{table}
Each row in the table corresponds to a single recorded data point, in this case representing the execution of an activity, in the context of an instance of the process.
We are able to relate the different data points by means of the \emph{Case id} column, which allows us to identify the underlying process instance.
Observe that, the data elements describe several data attributes, e.g., the timestamp of the activity, the resource that executed the activity, etc., illustrating the practical relevance of this type of data.
However, for the purpose of this paper, we primarily focus on sequences of data items that we are able to represent as a single symbol.
For example, when using the short-hand activity notation of the activities listed in \autoref{tab:example_log}, we obtain the sequence $\langle a,d,b,e,h \rangle$, for the process instance represented by case-id $1337$.\footnote{Note that, in the general sense, we are always able to map complex alpha-numerical string data into single characters.}

In the remainder, we let $\Sigma$ to denote the \emph{set of symbols}. 
In the context of this paper, a \emph{sequence database} (often called \emph{event log} in process mining) is defined as a finite multiset of sequences, $\mathit{L}{\in}\mathcal{B}({\Sigma^*})$.
A \emph{word} is a sequence $\sigma{\in}\mathit{L}$, i.e. a sequence of symbols in a sequence database.
For example, the sequence database $\mathit{L}{=}[\langle a,b,c\rangle^2,\langle b,a,c\rangle^3]$ consists of two occurrences of the word $\langle a,b,c\rangle$ and three occurrences of word $\langle b,a,c\rangle$.




\subsection{Process Models and Alignments}
In this section, we introduce process mining oriented concepts that we will apply in \autoref{sec:petri_net_predictions} for next-element prediction on the basis of process modeling formalisms.
In particular, we focus on two commonly used process modeling formalisms, i.e. \emph{automata} and \emph{Petri nets}.
Furthermore, we introduce the concept of an alignment, which allows us to explicitly explain a data sequence in the context of the behavior described by a Petri net.

\subsubsection{Probabilistic Automata}
Automata are a well-studied concept in process mining.\footnote{In some work in process mining, the automata used are alternatively called \emph{Transition Systems}.}
An automaton allows us to describe the possible states of a system, as well as the ways in which the system is able to change its state.
It furthermore describes an initial state and set of final states.
Automata have been used for several different purposes in the context of process mining.
In some cases they are directly used as a model of the process~\cite{DBLP:journals/is/BoltLA18,DBLP:conf/bpm/EckSA16,DBLP:conf/wecwis/EckSA17}.
In other cases, they are used as an intermediate representation of the sequence database, which is subsequently translated into another process model~\cite{DBLP:journals/sosym/AalstRVDKG10} in a more interpretable notation, such as the Petri net notation that we will introduce shortly hereafter.
Other applications of automata in process mining include noise filtering~\cite{DBLP:conf/caise/ZelstSOCR18} and several prediction tasks in business processes~\cite{Aalst2011,DBLP:conf/bpm/SchonenbergWDA08}.
In the context of this paper, we explicitly use the notion of probabilistic automata, which not only allow us to inspect a state of the system and its possible actions, it also allows us to quantify the probabilities of these actions.

\begin{definition}[Probabilistic Automaton]
\label{def:prob_aut}
Let $Q$ denote a set of states and let $\Sigma$ denote a finite set of symbols.
Furthermore, let $\delta \colon Q \times \Sigma \to \mathcal{P}(Q)$ denote a transition function and let $\gamma \colon Q \times \Sigma \times Q \to [0,1]$ denote a transition probability function.
Finally, let $q_0 \in Q$ represent the initial state and let $F \subset Q$ denote the set of {accepting states}.
Tuple $\mathit{PA} = (Q,\Sigma,\delta,\gamma,q_0,F)$ is a \emph{probabilistic automaton}, if and only if:
\begin{equation}
	\label{eq:prob_aut_rel_elem_prob}
	\forall q,q' \in Q, a \in \Sigma \left( q' \in \delta(q,a) \Leftrightarrow \gamma(q,a,q') > 0 \right)
\end{equation}
\begin{equation}
	\label{eq:prob_aut_sum_1}
	\forall q \in Q \left(  \left(\sum\limits_{a \in \Sigma} \sum\limits_{\{q' \in Q \mid q' \in \delta(q,a)\}} \gamma(q,a,q') \right) = 1  \right)
\end{equation}
For a given state $q \in Q$ and label $a \in \Sigma$, we denote the conditional probability of observing label $a$, whilst being in state $q$, as $P(a \mid q)$, where:
\begin{equation}
	P(a \mid q) = \sum\limits_{\{q' \in Q \mid q' \in \delta(q,a)\}} \gamma(q,a,q')
\end{equation}
\end{definition}
Observe that, \autoref{eq:prob_aut_rel_elem_prob} of \autoref{def:prob_aut} states that, if there exists a transition with label $a$, from state $q$ to state $q'$, the corresponding transition probability is non-zero.
Requirement \autoref{eq:prob_aut_sum_1} states that the sum of the probabilities of all outgoing transitions of a given state should equal $1$.

\begin{figure}[tb]
	\centering
		\begin{tikzpicture}[->,>=stealth',shorten >=1pt,auto,node distance=2cm,semithick]			
		\node[initial,state] (q1)                    {$q_1$};
		\node[state]    (q2) [right of=q1] {$q_2$};
		\node[state]	(q3) [above right of = q2] {$q_3$};
		\node[state]	(q4) [below right of = q2] {$q_4$};
		\node[state]	(q5) [below right of = q3] {$q_5$};
		\node[state]	(q6) [right of =  q5] {$q_6$};
		\node[state,accepting]	(q7) [right of =  q6] {$q_7$};

		\path (q1) edge  node {$a(1)$} (q2);
		\path (q2) edge [bend left = 15] node {$b(\frac{1}{4})$} (q3);
		\path (q2) edge [bend right = 15] node[label= below right:{$c(\frac{1}{4})$}] {} (q3);
		\path (q2) edge node[label=below left:{$d(\frac{1}{2})$}] {} (q4);
		\path (q3) edge node {$d(1)$} (q5);
		\path (q4) edge [bend left = 15] node {$b(\frac{1}{2})$} (q5);
		\path (q4) edge [bend right = 15] node[label= below right:{$c(\frac{1}{2})$}] {} (q5);
		\path (q5) edge node[] {$e(1)$} (q6);
		\path (q6) edge [out=-95,in=-125,distance=2.5cm] node {$f(\frac{1}{3})$} (q2);
		\path (q6) edge [bend left = 15] node {$g(\frac{1}{3})$} (q7);
		\path (q6) edge [bend right = 15] node[label= below:{$h(\frac{1}{3})$}] {} (q7);				
		\end{tikzpicture}
	\caption{An example probabilistic automaton. 
		The corresponding probabilities, i.e. $\gamma(q,a,q')$, are listed next to the transition names, e.g. $b(\frac{1}{2})$.}    
	\label{fig:automaton}
\end{figure}
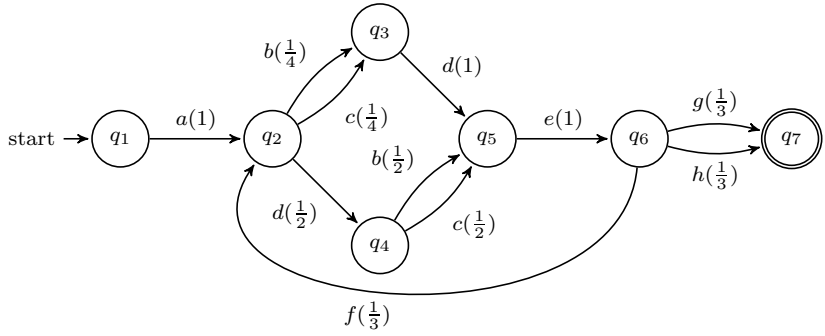

\subsubsection{Petri Nets}
\label{sec:petri}
Petri nets are a commonly used process model formalism to represent processes in process mining.   
Petri nets have several desirable properties that underly their popularity. First, they allow for explicit modeling of concurrent (i.e., parallel) behavior in a relatively compact manner. Secondly, most high-level, business-oriented, process modeling notations, e.g. BPMN~\cite{bpmn_2011}, are often translatable into Petri nets. And finally, the formal properties of Petri nets are well studies (e.g., see ~\cite{Esparza1994}). We now proceed to give a concise introduction of Petri nets and we refer to~\cite{Murata1989} for a more thorough and complete introduction.

\begin{figure}[tb]
	\resizebox{\textwidth}{!}{
		\begin{tikzpicture}
		[   node distance=1.70cm,
		on grid,>=stealth',
		bend angle=30,
		auto,
		every place/.style= {minimum size=6mm},
		every transition/.style = {minimum size = 6mm}
		]
		
		
		\node [place, tokens = 1] (start) [label=below:$p_1$]{};
		
		\node [transition] (t1) [label=below:$t_1$, label=center:$a$, label={[align=left]\scriptsize register request\ \ \ \ \ \ \ \ \ \ \ }, right = of start] {}
		edge [pre] node[auto] {} (start);
		
		\node [place] (p1) [label=above:$p_2$, above right = of t1] {}
		edge [pre] node[auto] {} (t1);
		
		\node [transition] (t2) [label=above:$t_3$, label=center:$c$, label=below:{\scriptsize \ \ \ examine thoroughly}, right = of p1] {}
		edge [pre] node[auto] {} (p1);
		
		\node [transition] (t22) [label=below:$t_2$, label=center:$b$, label={[align=left]\scriptsize \ \ \ \  examine casually}, above of = t2] {}
		edge [pre] node[auto] {} (p1);
		
		\node [place] (p2) [label=below:$p_3\ \ $, below right = of t1] {}
		edge [pre] node[auto] {} (t1);	
		
		\node [transition] (t3) [label=below:$\ \ \ \ \ \ \ t_4$, label=center:$d$, label=above:{\scriptsize check ticket}, right = of p2] {}
		edge [pre] node[auto] {} (p2);
				
		\node [place] (p3) [label=below:$p_4$, right = of t2] {}
		edge [pre] node[auto] {} (t2)
		edge [pre] node[auto] {} (t22);
		
		\node [place] (p4) [label=below:$p_5$, right = of t3] {}
		edge [pre] node[auto] {} (t3);
		
		\node [transition] (t5) [label=below:$t_6$, label=center:$e$, label=above:{\scriptsize \ \ decide}, above right = of p4] {}
		edge [pre] node[auto] {} (p3)
		edge [pre] node[auto] {} (p4);
		
		\node [transition] (t4) [fill=gray, minimum width=2mm, label=below:$t_5$, above = of t5] {}
		edge [pre] node[auto] {} (p3)
		edge [pre] node[auto] {} (p4);	
		
		\node [place] (p5) [label=below:$\ \ \ p_6$, right = of t5] {}
		edge [pre] node[auto] {} (t5)
		edge [pre] node[auto] {} (t4);
		
		\node [transition] (t6) [label=above:$t_7$, label=center:$f$, label=below:{\scriptsize\ \ \ \ \ \ \ \text{restart}} , right = of p4] {}
		edge [pre, bend right] node[auto] {} (p5)
		edge [post, bend left = 35] node[auto] {} (p2)
		edge [post, bend left = 100] node[auto] {} (p1);
		
		\node [transition] (t7) [label=below:$t_8$, label=center:$g$, label={[align=center]\scriptsize pay \\ compensation}, above right = of p5] {}
		edge [pre] node[auto] {} (p5);
		
		\node [transition] (t8) [label=above:$t_9$, label=center:$h$, label=below:{\scriptsize reject request}, below right = of p5] {}
		edge [pre] node[auto] {} (p5);
		
		\node [place] (end) [label=below:$p_7$, below right = of t7, pattern=custom north west lines,hatchspread=1.5pt,hatchthickness=0.3pt,hatchcolor=gray, label=center:{1}] {}
		edge [pre] node[auto] {} (t7)
		edge [pre] node[auto] {} (t8);		
		\end{tikzpicture}
	}
	\caption{An example accepting Petri net $\APN$, adopted from~\cite{Aalst2016}.
		The initial marking of the $\APN$ is $[p_1]$, the final marking is $[p_7]$}
	\label{fig:pt_net_example}
\end{figure}
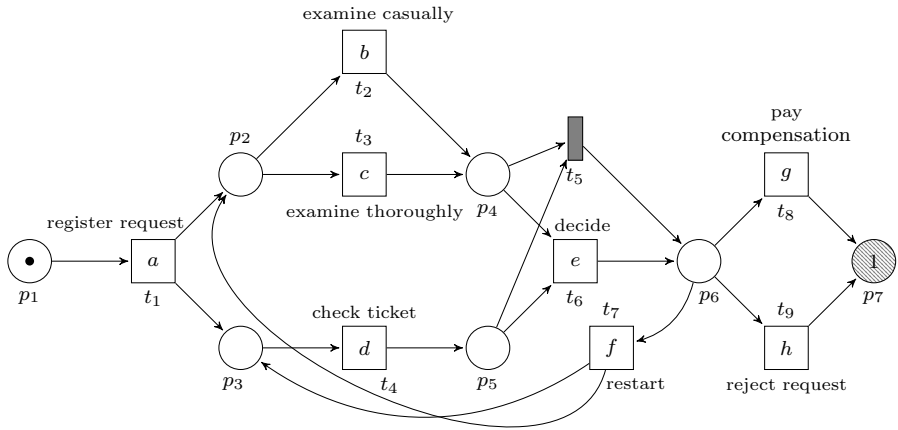

A Petri net is a directed bipartite graph consisting of places (depicted as circles) and transitions (depicted as rectangles), connected by arcs. 
The transitions allow us to describe the possible activities/symbols of the process, whereas the places represent the enabling conditions of transitions. 
The label of a transition indicates the symbol that the transition represents. 
Unlabelled transitions ($\tau$-transitions) represent invisible transitions (depicted as grey rectangles), which are used for routing purposes and are unobservable. 
As an example of a Petri net, consider \autoref{fig:pt_net_example}, which contains $7$ places and $9$ transitions.
The symbol corresponding to transition $t_1$ is symbol $a$, whereas transition $t_5$ is unobservable.

\begin{definition}[Labelled Petri net]
	\label{def:lpn}
	Let $\Sigma$ denote the universe of labels and let $\tau \notin \Sigma$.
	A \emph{labelled Petri net} $N{=}{(P,T,F,\ell)}$ is a tuple where $P$ is a finite set of places, $T$ is a finite set of transitions with $P{\cap}T{=}\emptyset$,  $F{\subseteq}(P {\times}T){\cup}(T{\times}P)$ describes the Petri net flow relation (graphically represented by means of arcs), and $\ell{\colon}T{\to}\Sigma \cup \{\tau\}$ is a labelling function that assigns a label to a transition, or leaves it unlabelled (the $\tau$-labelled transitions).
\end{definition}

We write $\bullet{x}$ and $x\bullet$ for the input and output nodes of $x\in P \cup T$ (according to $F$), e.g. in \autoref{fig:pt_net_example} we have $p_1\bullet = \{t_1\}$ and $\bullet t_3 = \{p_4\}$.
If there is no label known for $t\in T$, we have $\ell(t)=\tau$, i.e. $\tau \notin \Sigma$. 
The state of a Petri net is defined by its \emph{marking} $m{\in} \mathcal{B}(P)$ being a multiset of places. 
A marking is graphically denoted by putting $m(p)$ tokens in each place $p{\in}P$. 
For example, consider the marking of the example Petri net in \autoref{fig:pt_net_example}, i.e. $[p_1]$, represented by the black dot drawn in $p_1$.
State changes of a Petri net occur through \emph{transition firings}. 
A transition $t$ is enabled (can fire) in a given marking $m$ if each input place $p{\in}{\bullet}t$ contains at least one token. 
Once $t$ fires, one token is removed from each input place $p{\in}{\bullet} t$ and one token is added to each output place $p'{\in}t \bullet$, leading to a new marking $m'{=}m{-}\bullet\!{t}+t\bullet$. 
Firing a transition $t$ in marking $m$, yielding marking $m'$, is denoted as step $m {\step{t}} m'$. 
Several subsequent firing steps are lifted to sequences of firing  enabled transitions, written $m {\step{\sigma}} m'$ for $\sigma {\in}T^*$, and are referred to as a \emph{firing sequence}.

Defining an \emph{initial} and \emph{final} markings allow us to define the \emph{language} that is accepted by a Petri net as a set of finite sequences of symbols. 
To this end, we define the notion of an \emph{accepting Petri net}, i.e. a Petri net including an explicit initial and final marking.
\begin{definition}[Accepting Petri Net]
An \emph{accepting Petri net} is a triplet $\APN=(N,m_0,m_f)$, where $N=( P,T,F,\ell)$ is a labelled Petri net, $m_0{\in}\mathcal{B}(P)$ is its initial marking, and $m_f{\in}\mathcal{B}(P)$ its final marking. 
A sequence $\sigma{\in}\Sigma^*$ is a \emph{word} of an accepting Petri net $\APN$ if there exists a firing sequence $m_0{\step{\sigma}}m_f$, $\sigma{\in}T^*$ and $\ell(\sigma)_{\downarrow_{\Sigma}}{=}\sigma$.
\end{definition}

In this paper, we visualize the places that belong to the initial marking, by means of drawing the appropriate amount of tokens, e.g. $p_1$ in \autoref{fig:pt_net_example}. 
Places belonging to the final marking are marked as 
$\begin{tikzpicture}
[node distance=1.4cm,
on grid,>=stealth',
bend angle=20,
auto,
every place/.style= {minimum size=3.5mm},
]
\node [place,pattern=custom north west lines,hatchspread=1.5pt,hatchthickness=0.3pt,hatchcolor=gray, label=center:{1}] {};
\end{tikzpicture}$, i.e. having a grey diagonal pattern and indicating the number tokens required in the final marking.

Observe that the accepting Petri net depicted in \autoref{fig:pt_net_example}, describes similar behavior w.r.t. the sequence database listed in \autoref{tab:example_log}.
First, a request is registered, after which we observe a parallel branch describing that the ticket needs to be checked and examined (either casually or thoroughly).
After this, a decision is made.
In this model, the decision symbol is optional, e.g. given that the price of the ticket was lower than $\$ 10$ and the examination is positive, the compensation is directly granted.
Finally, we are able to restart the checking/examination, pay the compensation or reject the request.

The \emph{language} $\Lan(\APN)$ of an accepting Petri net is defined as the set of all its words, i.e. $\Lan(\APN)=\{ \sigma \in \Sigma^* \mid \exists \sigma \in T^*(\ell(\sigma) = \sigma \wedge m_0{\step{\sigma}}m_f)\}$, which is potentially of infinite size, i.e. when $\APN$ contains loops. 
While we define the language for accepting Petri nets, in theory, 
$\Lan(M)$ can be defined for any process model $M$ with formal semantics. 
We denote the universe of process models as $\mathcal{M}$ and assume that for each $M{\in}\mathcal{M}$, $\Lan(M)\subseteq\Sigma^*$ is defined.

A process discovery method is a function $\mathit{PD}:\mathcal{B}({\Sigma^*})\rightarrow\mathcal{M}$ that produces a process model from an sequence database. 
The discovered process model should cover as much as possible the behavior observed in the sequence database (a property called \emph{fitness}) while it should not allow for too much behavior that is not observed in the sequence database (called \emph{precision}). 
For an sequence database $\mathit{L}$, $\tilde{\mathit{L}}{=}\{\sigma{\in}\Sigma^*|\mathit{L}(\sigma){>}0\}$ is also referred to as the \emph{word set} of $\mathit{L}$. 
For example, for sequence database $\mathit{L}{=}[\langle a,b,c\rangle^2,\langle b,a,c\rangle^3]$, $\tilde{\mathit{L}}{=}\{\langle a,b,c\rangle\langle b,a,c\rangle\}$.
For an sequence database $\mathit{L}$ and a process model $M$, we say that $\mathit{L}$ is \emph{fitting} on $M$ if $\tilde{\mathit{L}}{\subseteq}\Lan(M)$. 
\emph{Precision} is related to the behavior that is allowed by a model $M$ that was not observed in sequence database $\mathit{L}$, i.e. $\Lan(M){\setminus}\tilde{\mathit{L}}$.

\subsubsection{Alignments}
\label{ssec:alignments}
Revisit the the accepting Petri net depicted in \autoref{fig:pt_net_example}.
Furthermore, assume we are given the word $\langle a,b,e,e,g\rangle$ (using short-hand activity notation).
Clearly, the given word is not in correspondence with the process model depicted in \autoref{fig:pt_net_example}, i.e. it is not in the language of the model.
The word is missing the label $d$, and, symbol $e$ is (unnecessarily) duplicated.

\emph{Alignments} allow us to compute and quantify to what degree observed behavior corresponds to a given reference model.
Consider \autoref{fig:alignment_example}, in which we present two different alignments of the word $\langle a,b,e,e,g\rangle$ w.r.t. the accepting Petri net depicted in \autoref{fig:pt_net_example}.
\begin{figure}[tb]
	{
		\resizebox{\textwidth}{!}{
			\begin{tabular}{l||c|c|c|c|c|c|}
				\multirow{2}{*}{$\alpha_1:$} & $a$ & $b$ & $\gg$ & $e$ & $e$ & $g$\\
				\cline{2-7}
				& $t_1$ & $t_2$ & $t_4$ & $t_6$ & $\gg$ & $t_8$
			\end{tabular}~\hfill
			\begin{tabular}{l||c|c|c|c|c|c|c|c|c|}
				\multirow{2}{*}{$\alpha_2:$} & $a$ & $b$ & $\gg$ & $e$ & $\gg$ & $\gg$ & $\gg$ & $e$ & $g$\\
				\cline{2-10}
				& $\gg$ & $t_1$ & $t_4$ & $t_6$ & $t_7$ & $t_3$ & $t_4$ & $t_6$ & $t_8$
			\end{tabular}
		}
	}
	\caption{Two example alignments, i.e. $\alpha_1$ and $\alpha_2$, for the word $\langle a,b,e,e,g \rangle$ w.r.t. the accepting Petri net presented in \autoref{fig:pt_net_example}.
	We prefer $\alpha_1$ over $\alpha_2$, since it explains the given behavior using less deviations. }
	\label{fig:alignment_example}
\end{figure}
Alignments are sequences of pairs, e.g. $\alpha_1 = \langle (a, t_1), (b,t_2), ..., (g, t_8) \rangle$.
Each pair within an alignment is referred to as a \emph{move}.
The first element of a move refers to a symbol in the given word whereas the second element refers to a transition.
The goal is to create pairs of the form $(a, t)$ s.t. $\ell(t) = a$, i.e. the execution of a transition in the model corresponds (in terms of its label) with the observed symbol, referred to as a \emph{synchronous move}.
In some cases it is not possible to construct a move of the form $(a, t)$ s.t. $\ell(t) = a$, e.g. $(\gg,t_4)$ and $(e,\gg)$ in $\alpha_1$. Moves of the form $(\gg,t)$ represent that we did not observe a symbol in the event data, that was expected according to the model, referred to as \emph{model moves}.\footnote{Observe that, when $\ell(t)=\tau$, we are never able to observe a corresponding symbol, and thus, even though such transition always is of the form $(\gg,t)$, we often consider such moves as being synchronous moves.}
Moves of the form $(a,\gg)$ represent the opposite, i.e. we observe a symbol that was not supposed to be observed according to the model, referred to as \emph{log moves}.

The sequence of symbol labels in the alignment needs to equal the input word (when ignoring the $\gg$-symbols).
The sequence of transitions in the alignment needs to correspond to a $\sigma \in T^*$ s.t., given the designated initial marking $m_i$ and final marking $m_f$ of the process model, we have $m_i \xrightarrow{\sigma} m_f$ (again ignoring the $\gg$-symbols).
For the Petri net presented in \autoref{fig:pt_net_example}, we have $m_i = [p_1]$ and $m_f = [p_7]$.

As the alignments presented in \autoref{fig:alignment_example} signify, several alignments exist for a given word and process model.
In general, we are interested in finding an alignment that minimizes the number of log- and/or model moves (such an optimal alignment is not necessarily unique).

In the purpose of making next-element predictions with process models, we are interested in a particular type of alignments called \emph{prefix-alignments}. 
We will show in~\autoref{sec:petri_net_predictions} how to leverage prefix-alignments to make predictions with process models.
The main difference w.r.t. the conventional alignments described above relates to the ``model-part'' of the alignments.
For a prefix-alignment, the model-part of the alignment does not have to finish in the final marking $m_f$, but instead, it just has to finish in some marking $m$ from which $m_f$ is still reachable.
More formally:
\begin{center}
	\emph{Given the designated initial marking $m_i$ and final marking $m_f$ of the process model, a prefix-alignment corresponds to a model run $m_i\xrightarrow{\sigma} m$, for some marking $m\in \mathcal{B}(P)$ such that there exists a $\sigma'\in  T^*$ for which $m \xrightarrow{\sigma'} m_f$.}
\end{center}

For example, given word $\langle a,b \rangle$, a conventional alignment of the word w.r.t the model in \autoref{fig:pt_net_example}, is $\langle (a,t_1), (b,t_2), (\gg,t_4), (\gg,t_5), (\gg,t_8) \rangle$, whereas a prefix-alignment is simply $\langle (a,t_1), (b,t_2) \rangle$, i.e. we have $[p_1] \xrightarrow{\langle t_1,t_2\rangle}[p_3,p_4]\xrightarrow{t_4,t_5,t_8}[p_7]$.

It follows from the definition of prefix-alignments that $m_i\xrightarrow{\sigma}m$ always yields a prefix of a word from the process model (ignoring the $\gg$-symbols), i.e., there exists a $\sigma'\in\Lan(M)$ such that $\ell(\sigma)$ is a prefix of $\sigma'$. This contrasts conventional alignments, which always yield exactly a word from the model language instead of a prefix from it.

A formal definition of the algorithm to compute alignments and prefix-alignments for arbitrary words of process behavior and process models is out of the scope of this paper and we refer to~\cite{Adriansyah2014, DBLP:conf/apn/ZelstBD17}.

\section{Next Element Prediction Methods}
\label{sec:prediction}
In this section, we present several methods to predict the probability distribution over the next element following a given incomplete sequence. These methods originate from different research fields. We start by introducing several sequence models from the machine learning community: neural networks in~\autoref{ssec:black-box_sequence_models} and Markov models in~\autoref{ssec:methods:markov}. We continue this section by introducing grammar inference in~\autoref{ssec:grammar-inference}. In~\autoref{ssec:methods:automaton} we introduce some automaton-based techniques for next-element prediction that originate from the process mining community. Another class of methods from the process mining community is presented in~\autoref{sec:petri_net_predictions}, where we present how Petri nets can be used as a sequence model, thereby enabling the use of process discovery approaches as interpretable sequence models. We conclude this section on sequence modeling methods in~\autoref{ssec:sequence-model-sumamry}, where we discuss similarities and differences between the techniques presented in this section. 

\subsection{Neural Networks \& Recurrent Neural Networks}
\label{ssec:black-box_sequence_models}
A neural network consists of one layer of \emph{input units}, one layer of output units, and in-between are one or more layers that are referred to as \emph{hidden units}. The outputs of the input units form the inputs for the units of the first \emph{hidden layer} (i.e. the first layer of hidden units), and the outputs of the units of each hidden layer form the input for each subsequent hidden layer. The outputs of the last hidden layer form the input for the output layer. The output of each unit is a function over the weighted sum of its inputs. The weights of this weighted sum performed in each unit are optimized iteratively by applying the current weights to some training sequences and back-propagating the partial derivative of the weights with respect to the error back through the network and adjusting the weights accordingly. Recurrent Neural Networks (RNNs) are a special type of neural networks where the connections between neurons form a directed cycle.

\begin{figure}[t]
	\centering
	\tikzset{
		neuron/.style={ 
			circle,draw,thick, 
			inner sep=0pt, 
			minimum size=3.5em, 
			node distance=1ex and 2em, 
		},
		group/.style={ 
			rectangle,draw,thick, 
			inner sep=0pt, 
		},
		io/.style={ 
			neuron, 
			fill=gray!15, 
		},
		conn/.style={ 
			-{Straight Barb[angle=60:2pt 3]}, 
			thick, 
		},
	}
	\resizebox{0.8\linewidth}{!}{
		\subfloat[]{
			\begin{tikzpicture}
			\node[neuron] (ht1) {$s$};
			\node[io,above=2em of ht1] (yt) {$o$};
			\node[io,below=2em of ht1] (vt) {$x$};
			\draw[conn] (ht1) -- (yt) node[midway, right] {$V$};
			\draw[conn] (vt) -- (ht1) node[midway, right] {$U$};
			\draw[conn] (ht1) to [loop right] node[midway,right] {$W$} (ht1);
			\end{tikzpicture}
		}
		\qquad
		\subfloat[]{
			\begin{tikzpicture}
			\foreach \jlabel [count=\j, evaluate={\k=int(mod(\j-1,1)); \jj=int(\j-1);}]
			in {t-2, t-1, t, t+1, t+2}{
				\foreach \ilabel [count=\i] in {1}
				\node [neuron] at (2*\j, 1-\i) (h-1-\j){$s_{\jlabel}$};
				\ifcase\k
				\node [io, above=2em of h-1-\j] (y-\j) {$o_{\jlabel}$};
				\node [io, below=2em of h-1-\j] (v-\j) {$x_{\jlabel}$};
				\draw [conn] (v-\j) -- (h-1-\j) node[midway, right] {$U$};
				\draw [conn] (h-1-\j) -- (y-\j) node[midway, right] {$V$};
				\fi
				\ifnum\j>1
				\draw [conn] (h-1-\jj.east) -- (h-1-\j.west) node[midway, below]{$W$};
				\fi
			} 
			\node [left=of h-1-1] {\ldots};
			\node [right=of h-1-5] {\ldots};
			\end{tikzpicture}
	}}
	\caption{\emph{(a)} A simple recurrent neural network consisting of a single hidden layer, and \emph{(b)} the recurrent neural network unfolded over time.}
	\label{fig:RNNexample}
\end{figure}
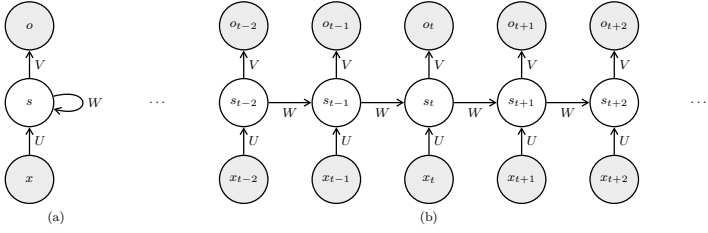
RNNs can be unfolded, as shown in \autoref{fig:RNNexample}. Each step in the unfolding is referred to as a time step, where $x_t$ is the input at time step $t$. RNNs can take an arbitrary length sequence as input, by providing the RNN a feature representation of one element of the sequence at each time step. $s_t$ is the hidden state at time step $t$ and contains information extracted from all time steps up to $t$. The hidden state $s$ is updated with information of the new input $x_t$ after each time step: $s_t = f(Ux_t+Ws_{t-1})$, where $U$ and $W$ are vectors of weights over the new inputs and the hidden state respectively. In practice, either the hyperbolic tangent or the logistic function is generally used for function $f$, which is referred to as the activation function. The logistic function is defined as: $\mathit{sigmoid}(x)=\frac{1}{1+\mathit{exp}(-x)}$. In neural network literature, the sigmoid function is often represented by $\sigma$, however, to avoid confusion with sequences, we fully write $\mathit{sigmoid}$. $o_t$ is the output at step $t$.

\subsubsection{Long Short-Term Memory for Sequence modeling}
A Long Short-Term Memory (LSTM) model \cite{Hochreiter1997} is a special Recurrent Neural Network architecture that has powerful modeling capabilities for long-term dependencies. The main distinction between a regular RNN and an LSTM is that the latter has a more complex memory cell $C_t$ replacing $s_t$. Where the value of state $s_t$ in an RNN is the result of a function over the weighted average over $s_{t-1}$ and $x_t$, the LSTM state $C_t$ is accessed, written, and cleared through controlling gates, respectively $o_t$, $i_t$, and $f_t$. Information on a new input will be accumulated to the memory cell if $i_t$ is activated. Additionally, the previous memory cell value $C_{t-1}$ can be ``forgotten'' if $f_t$ is activated. The information of $C_t$ will be propagated to the output $h_t$ based on the activation of output gate $o_t$. Combined, the LSTM model can be described by the following formulas:
\begin{align*}
	f_t&=\mathit{sigmoid}(W_f\cdot[h_{t-1},x_t]+b_f) & i_t&=\mathit{sigmoid}(W_i\cdot[h_{t-1},x_t]+b_i)\\
	\tilde{C}_t&=\mathit{tanh}(W_c\cdot[h_{t-1},x_t]+b_{C}) & C_t&=f_t*C_{t-1}+i_i*\tilde{C}_t\\
	o_t&=\mathit{sigmoid}(W_o[h_{t-1},x_t]+b_o) & h_t&=o_t*\mathit{tanh}(C_t)
\end{align*}
In these formulas all $W$ variables are weights and $b$ variables are biases and both are learned during the training phase.

\subsubsection{Gated Recurrent Units}
Gated Recurrent Units (GRU) were proposed by Cho et al.~\cite{Cho2014} as a simpler alternative to the LSTM architecture. In comparison to LSTMs, GRUs do not keep a separate memory cell and instead merge the cell state $C_t$ and hidden state $h_t$. Furthermore, a GRU combines the input gate $i_t$ and the forget gate $f_t$ into a single \emph{update gate}. While the LSTMs and GRUs are identical in the class of functions that they can learn, GRUs are simpler in the sense that they have fewer model parameters. Empirically, GRUs have been found to outperform LSTMs on several sequence prediction tasks~\cite{Chung2014}.

\subsection{Markov Models}
\label{ssec:methods:markov}
A Markov model is a stochastic model commonly used in probability theory in order to model randomly changing systems. The model makes the assumption of the \emph{Markov property}, i.e. that the future state of the system only depends on the present state. 
The simplest type of Markov models are \emph{Markov chains}, which can be used when the states of the system are fully observable. 
The parameters of a Markov chain consist of a matrix of \emph{transition probabilities} expressing the likelihood of transitioning from any given state to any other state. In a 1st-order Markov chain, the state represents the last observed symbol of in the sequence. In Markov chains of higher order, the state represents a longer window of observed symbols, i.e., in a $k^{\mathit{th}}$-order Markov model, the state represents the last $k$ symbols. 

A sequence model called \emph{all k-order Markov models} (AKOM)~\cite{Pitkow1999} is an extension of Markov chains that fits all models up to an order $k$ to the training sequences. When making a prediction (i.e. estimating the transition probability from a given state in a test sequence), AKOM uses the Markov model with the highest $k$ that has a state that matches the test sequence.

\emph{Hidden Markov models} (HMM) are a type of Markov model where the states represent \emph{latent variables} that do not represent some window over the sequences directly but instead represent some unobserved property that is inferred from the sequence. An HMM assumes that the system can be described in terms of a number of \emph{hidden states} that are not directly observable in the data. Therefore, the next observation in the sequence depends not only on the most recent observation, but also on the likelihood of being in a particular hidden state. The parameters of an HMM include the transition probabilities, i.e. transitioning from a hidden state to another hidden state, and the \emph{emission probabilities}, expressing the likelihood of a particular observation while being in a given hidden state. The transition probabilities between states and the emission probabilities from states to symbols are learned from a training dataset using the Baum-Welch algorithm. When predicting the next symbol for a sequence with an HMM, one can either 1) apply the well-known Viterbi algorithm to extract the most likely sequence of hidden states for the given sequence and make the prediction according to the emission probabilities of the final hidden state, or 2)  apply the forward algorithm to obtain a probability distribution over the likelihoods of the hidden states given the sequence and make the prediction by weighting the emission probabilities of those states by their hidden state likelihood.

\subsection{Grammar Inference}
\label{ssec:grammar-inference}
The research field of \emph{grammar inference}, also called \emph{grammar induction}, is concerned with learning a grammar that describes a language based on a collection of positive examples of elements from this language. Observe that the grammar inference field closely links to the area of process discovery, where the dataset of positive examples are called an event log and the language that is learned is represented as a process model instead of a formal grammar. A variety of grammar inference techniques exists. \emph{Automaton learning} techniques focus on learning a \emph{deterministic finite automaton} that describes the language and can be used when the language is assumed to belong to the class of \emph{regular languages}. Other grammar inference techniques assume the language to belong to the class of \emph{context-free languages} and focus on extracting a context-free language in extended Backus-Naur form~\cite{Wirth1977}. We refer to~\cite{Higuera2005} and~~\cite{Higuera2010} for an extensive overview of the grammar inference field.

The grammar inference field puts special emphasis on formal analysis of the learnability of languages. Early work by Gold~\cite{Gold1978}, who proved that the problem of finding the smallest DFA consistent with a given set of strings is NP-hard, plays a central role in the grammar inference field. Angluin~\cite{Angluin1987} proposed an \emph{active learning} setting of grammar inference where the algorithm does not learn the grammar from a fixed set of positive samples, but instead iteratively queries the world with strings for which it wants to obtain whether or not this string is or is not part of the language. Angluin~\cite{Angluin1987} proved that regular languages can be identified in a polynomial amount of queries using active learning and proposed an called L* that is able to do so. In this paper, we assume a dataset of positive examples of sequences from a language to be given and therefore we leave active learning algorithms to grammar inference out of scope. 

Some grammar induction techniques focus on learning probabilistic grammars, i.e., they do not just specify which sequences are included in and which are excluded from the language, but they additionally specify the likelihoods of each word of the language.

Several competitions have been organized in the grammar inference field that aimed at benchmarking grammar inference methods and tools. The Abbadingo challenge (1998)~\cite{Lang1998} focused on learning deterministic finite state automata, Omphalos (2004)~\cite{Clark2007} focused on learning context-free grammars, and PAutomaC (2014)~\cite{Verwer2014} on learning probabilistic finite state machines. The \underline{S}equence \underline{P}red\underline{I}ction \underline{C}halleng\underline{E} (SPiCe)~\cite{Balle2017} was a recent challenge from the grammar inference field that defined the challenge task similar to the focus of this paper: predicting the next symbol in a sequence. The SPiCE competition used a well-known probabilistic automaton learning algorithm from the grammar inference field as a baseline method, which is called \emph{spectral learning}~\cite{Balle2014}.

A central concept in the spectral learning approach to grammar inference is the so-called Hankel matrix, which is a bi-infinite matrix in which the rows correspond to the prefixes of the sequences in the dataset and columns its suffixes. The value in a cell of the Hankel matrix represents the weight of the corresponding sequence in the corresponding weighted automaton. The rank of this matrix is the number of states in the minimal
weighted automaton. Balle et al.~\cite{Balle2014} showed that in this way the weighted automaton can be constructed through a rank factorization of the Hankel matrix. Spectral learning relies on constructing a finite sub-block approximation of the Hankel matrix and using a Singular Value Decomposition on the resulting matrix to obtain a rank factorization and thus a weighted automaton.

\subsection{Automaton-Based Prediction Techniques from the Process Mining Field}
\label{ssec:methods:automaton}
We now present a probabilistic automaton for next element prediction that is based \emph{abstraction functions}~\cite{DBLP:journals/sosym/AalstRVDKG10} that are frequently used in the process mining field. The automata that are constructed based on these abstractions have many applications in process mining. Van der Aalst et al.~\cite{DBLP:journals/sosym/AalstRVDKG10} first introduced them for the purpose of process discovery, where the obtained automata were ultimately transformed into a Petri net. More recently, these abstraction-based automata have been used to predict the remaining cycle time of a business process instance~\cite{Aalst2011} and, like here, to predict the next element of a sequence~\cite{Tax2017}.

\subsubsection{Training}
When constructing a probabilistic automaton for the purpose of next element prediction, we roughly perform two steps, i.e. 1) automaton construction and 2) transition probability computation.
In automaton construction, we conceptually transform each word into a sequence of automaton transitions, which we represent by tuples of the form $(q,a,q') \in Q \times \Sigma \times Q$.
The way in which we determine the states, based on the words in the sequence database, is strongly parametrized.


\begin{definition}[Word Abstraction Sequence]
\label{def:trace_abstractions}
For a given set of symbols $\Sigma$, sequence $\sigma \in \Sigma^*$ and $k \in \mathbb{N}^+$ we define the following abstraction functions:

\begin{itemize}
\item The \emph{sequence abstraction} function $\pi^k_{\texttt{seq}}{\colon}{\Sigma^*}{\to}{\left(\Sigma^* \times \Sigma \times \Sigma^* \right)^*}$:
\end{itemize}
\begin{equation}
\pi^k_{\texttt{seq}}(\sigma) = \left\langle \left(\epsilon, \sigma(1), \left\langle \sigma(1) \right\rangle\right),..., \left(\sigma_{(|\sigma|- k, |\sigma|-1)}, \sigma(|\sigma|), \sigma_{(|\sigma|- k + 1, |\sigma|)} \right)\right\rangle
\end{equation}

\begin{itemize}
	\item
The \emph{set abstraction} function $\pi^k_{\texttt{set}}{\colon}{\Sigma^*}{\to}{\left(\mathcal{P}(\Sigma) \times \Sigma \times \mathcal{P}(\Sigma) \right)^*}$:
\end{itemize}
\begin{equation}
	\pi^k_{\texttt{set}} = \left\langle \left(\emptyset, \sigma(1), \overline{\left\langle \sigma(1) \right\rangle}\right),..., \left(\overline{\sigma_{(|\sigma|- k, |\sigma|-1)}}, \sigma(|\sigma|), \overline{\sigma_{(|\sigma|- k + 1, |\sigma|)}} \right)\right\rangle
\end{equation}
\begin{itemize}
	\item
The \emph{multiset abstraction} function $\pi^k_{\texttt{mul}}{\colon}{\Sigma^*}{\to}{\left(\mathcal{B}(\Sigma) \times \Sigma \times \mathcal{B}(\Sigma) \right)^*}$:
\end{itemize}
\begin{equation}
\pi^k_{\texttt{mul}} = \left\langle \left([\ ], \sigma(1), \overrightarrow{\left\langle \sigma(1) \right\rangle}\right),..., \left(\overrightarrow{\sigma_{(|\sigma|- k, |\sigma|-1)}}, \sigma(|\sigma|), \overrightarrow{\sigma_{(|\sigma|- k + 1, |\sigma|)}} \right)\right\rangle
\end{equation}
\end{definition}

To give an example, consider applying the different abstractions as defined in \autoref{def:trace_abstractions}, on word $\langle a,b,b,c \rangle$, for $k=2$:
\begin{itemize}
	\item $\pi^2_{\texttt{seq}}(\langle a,b,b,c\rangle) = \left\langle \left(\epsilon,a,\langle a \rangle\right), \left(\langle a \rangle,b,\langle a,b \rangle\right), \left(\langle a,b \rangle,b,\langle b,b \rangle\right), \left(\langle b,b \rangle,c,\langle b,c \rangle\right) \right\rangle$
	\item $\pi^2_{\texttt{set}}(\langle a,b,b,c\rangle) = \left\langle \left(\emptyset,a,\{ a \}\right), \left(\{ a \},b,\{ a,b \}\right), \left(\{ a,b \},b,\{ b \}\right), \left(\{ b \},c,\{ b,c \}\right) \right\rangle$
	\item $\pi^2_{\texttt{mul}}(\langle a,b,b,c\rangle) = \left\langle \left([\ ],a,[ a ]\right), \left([ a ],b,[ a,b ]\right), \left([ a,b ],b,[ b^2 ]\right), \left([ b^2 ],c,[ b,c ]\right) \right\rangle$
\end{itemize}

\begin{figure}[tb]
	\centering
	\begin{tikzpicture}[->,>=stealth',shorten >=1pt,auto,node distance=2cm,semithick]			
	\node[initial,state,  minimum size=1cm] (q1)                    {$\emptyset$};
	\node[state, minimum size=1cm]    (q2) [right of=q1] {$\{a\}$};
	\node[state,  minimum size=1cm]    (q3) [right of=q2] {$\{a,b\}$};
	\node[state,  minimum size=1cm]    (q4) [right of=q3] {$\{b\}$};
	\node[state, accepting,  minimum size=1cm]    (q5) [right of=q4] {$\{b,c\}$};

	\path (q1) edge  node {$a (1)$} (q2);	
	\path (q2) edge  node {$b (1)$} (q3);
	\path (q3) edge  node {$b (1)$} (q4);
	\path (q4) edge  node {$c (1)$} (q5);
			
	\end{tikzpicture}
	\caption{An example probabilistic automaton, constructed using $\pi^2_{\texttt{set}}(\langle a,b,b,c\rangle)$.
	The automaton is based on only one word (with each state in the abstraction sequence unique), therefore, all transition probabilities are equal to $1$.}    
	\label{fig:learning_automaton}
\end{figure}
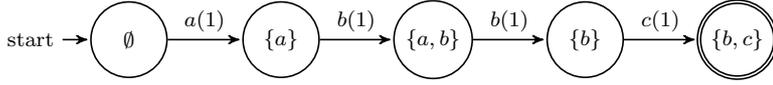

Hence, given an abstraction of choice, i.e. either \emph{set}, \emph{multiset} or \emph{sequence}, and a value for $k$, we are able to transform an sequence database into a multiset of abstraction sequences.
Translating such a multiset to an automaton is trivial, i.e. each first- and third element of the tuples present in the different abstraction sequences, as defined by the sequence database, represents a state.
The second element of each tuple present in the different abstraction sequences represents a transition.
Conceptually, when we have a tuple of the form $(q,a,q')$ present in the multiset of abstraction sequences of the sequence database, this implies that $q' \in \delta(q,a)$ in the corresponding resulting automaton.
Clearly, depending on the abstraction of choice, $\epsilon$, $\emptyset$ or $[\ ]$, is the initial state of the automaton.
The third argument of the last tuple in an abstraction sequence is an accepting state.
Reconsider $\pi^2_{\texttt{set}}(\langle a,b,b,c\rangle)$, and assume that $\langle a,b,b,c \rangle$ is the only word in a given sequence database.
The corresponding resulting automaton is depicted in \autoref{fig:learning_automaton}

As a final step, we learn the transition probabilities.
When constructing the automaton, we keep track of a counter $c \colon Q \times \Sigma \times Q \to \mathbb{N}^+$ that counts the number of times a certain abstraction sequence $(q,a,q')$ occurs, on the basis of the sequence database.
Thus, when we observe abstraction $(q,a,q')$, we increment the counter on the basis of triple $(q,a,q')$: $c(q,a,q') \gets c(q,a,q') + 1$.
After constructing the final automaton, the $\gamma$-values are computed as the empirical probability of occurrence, i.e.:
\begin{equation}
	\gamma(q,a,q') = \frac{c(q,a,q')}{\sum\limits_{a' \in \Sigma}\sum\limits_{q'' \in Q}c(q,a',q'')}
\end{equation}

\subsubsection{Prediction}
The distribution over the next symbol given an incomplete sequence is obtained by assessing the state in the probabilistic automaton that matches the sequence.
To this end, the same abstraction and value $k$ that were used in training is used in the prediction phase.
Given a prefix $\sigma' \in \Sigma^*$, we determine the current state by applying the abstraction of choice on top of $\mathit{tl}^k(\sigma')$.
Hence, in case of \emph{sequence}, it is simply $\mathit{tl}^k(\sigma')$, in case of \emph{set}, it is $\overline{\mathit{tl}^k(\sigma')}$, and in case of multiset, it is $\overrightarrow{\mathit{tl}^k(\sigma')}$.
For example, using a set abstraction, with $k=2$ on the prefix $\langle a, b, b\rangle$, yields, $\overline{\mathit{tl}^2(\langle a, b, b\rangle)} = \overline{\langle b,b \rangle} = \{b\}$.
Given the computed state $q$, we output each $P(a | q)$ as a prediction.
In the case of our example, c.f. \autoref{fig:learning_automaton}, we output observing label $c$ with probability $1$, and all other possible symbols with probability $0$.

Note that, by definition, the automata are able to generalize w.r.t. the training behavior, e.g. we obtain the exact same prediction on the basis of prefix $\langle a,b,b,a,b,c,b,b \rangle$ in combination with \autoref{fig:learning_automaton}, i.e. $\overline{\mathit{tl}^2(\langle a,b,b,a,b,c,b,b \rangle)} = \overline{\langle b,b \rangle} = \{b\}$.
However, observe that, given the automaton in \autoref{fig:learning_automaton}, for the prefix $\langle a,b,b,d \rangle$, we infer $\overline{\mathit{tl}^2(\langle a, b, b,d\rangle)} = \overline{\langle b,d \rangle} = \{b,d\}$, which does not exist in the automaton.
In such a case, we can not use the automaton for prediction, hence, we predict the global empirical probability distribution of the symbols on the basis of the training sequence database (total number of occurrences of a symbol divided by the total number of events in the log).
In our example, the automaton is based on an sequence database containing the single sequence $\langle a,b,b,c \rangle$, hence the global symbol distribution is $P(a)=P(c)=\frac{1}{4}$ and $P(b)=\frac{1}{2}$, which is the prediction for any state not present in the automaton.\looseness=-1

\subsection{Next-Element Prediction using Process Discovery}
\label{sec:petri_net_predictions}
We will now describe how process discovery methods can be used for the purpose of next-element prediction. First, we will describe how to make next-element predictions with any given Petri net in~\autoref{ssec:petrinet-prediction}. Secondly, in~\autoref{ssec:process-discovery-methods} we present how to actually obtain a Petri net based on training sequences, by giving a brief overview of the different process discovery algorithms that are considered in this paper.

\subsubsection{Next-Element Prediction using a Petri net}
\label{ssec:petrinet-prediction}
To use a Petri net $\APN$ as next-symbol predictor, we for now assume that we have a method $f$ to map a prefix to a marking $m$. We start with a training phase in which we deduce a probability distribution over symbols $\Sigma$ based on sequence database $L$ for each marking $m$ of $\APN$ that is reached when replaying $L$ (using $f$). After the training phase, when making a prediction with Petri net $\APN$ for a given prefix $\sigma$, we again map the prefix to a marking $m$ in $\APN$ and predict the next symbol according to the probability distribution that we had learned for $m$. We propose a two-step approach for the training-phase:
\begin{enumerate}
	\item We compute the most likely markings in Petri net $\APN$ for all observed prefixes of training sequences $L$.
    \item For each marking reached we compute a probability distribution describing the possible next elements $\Sigma \to [0,1]$.
\end{enumerate}

We describe these two steps in detail in the upcoming paragraphs.

\paragraph{Computing Prefix-Based Markings in a Petri net.}
To deduce what symbols are able to follow a prefix $\sigma$, using a Petri net $\APN$ as a sequence model, we obtain a marking of the model that is corresponds to firing the given prefix $\sigma$ in the Petri net.
A naive approach to this problem is to play the so-called ``token-game''.
In the token-game, starting from the initial marking, we simply fire enabled transitions in such a way that we obtain a firing sequence $\gamma{\in}T^*$ that projected on $\ell$ equals $\sigma$, and marks some arbitrary marking $m$ in $\APN$.
Such an approach works, as long as the observed prefix actually allows us to reach a marking $m$, i.e. the prefix should fit the model.
Furthermore, in case the Petri net contains multiple transitions describing the same label, i.e. $t,t'{\in}T$ with $t{\neq}t'$ and $\ell(t){=}\ell(t')$, such a strategy becomes more complex and potentially leads to ambiguous results. Therefore, we propose to calculate prefix-alignments, as introduced in~\autoref{ssec:alignments}, as they provide a natural solution to the two aforementioned problems.	

\begin{figure}[tb]
	\begin{center}
		\begin{tabular}{l||c|c|}
			\multirow{2}{*}{$\alpha_3:$} & $a$ & $b$\\
			\cline{2-3}
			& $t_1$ & $t_2$
		\end{tabular}~
		\begin{tabular}{l||c|c|c|c|}
			\multirow{2}{*}{$\alpha_4:$} & $a$ & $b$ & $\gg$ & $e$\\
			\cline{2-5}
			& $t_1$ & $t_2$ & $t_4$ & $t_6$
		\end{tabular}
	\end{center}
	\caption{Two optimal prefix-alignments ($\alpha_3$ and $\alpha_4$) w.r.t. the accepting Petri net of~\autoref{fig:pt_net_example} for the prefixes $\langle a,b \rangle$ and $\langle a,b,e\rangle$ respectively.}
	\label{fig:_pref_alignment_example}
\end{figure}

Consider \autoref{fig:_pref_alignment_example} in which we depict two prefix-alignments of two different prefixes of words, i.e. $\langle a,b \rangle$ and $\langle a,b,e \rangle$ in the context of the example accepting Petri net, depicted in \autoref{fig:pt_net_example}. It is easy to see that the word $\langle a,b \rangle$ complies with a prefix as described by the model.
Hence, the leftmost prefix-alignment describes that sequence $\langle t_1, t_2 \rangle$ is a firing sequence explaining the observed activity sequence $\langle a,b \rangle$.
Observe that, indeed, $\ell(t_1) = a$ and $\ell(t_2) = b$, and $[p_1]\xrightarrow{\langle t_1, t_2 \rangle}[p_3,p_4]$.

For sequence $\langle a,b,e \rangle$ however, we observe that it does not directly comply with a prefix as described by the model.
The rightmost prefix-alignment describes that sequence $\langle t_1, t_2,t_4,t_6 \rangle$ is a firing sequence that most accurately describes the observed behavioral sequence in terms of the model.
In this case, it states that the best possible way to explain the observed behavioral sequence, by means of assuming that label $d$ was not observed, whereas it was supposed to happen (as represented by the $(\gg,t_4)$) move.
Nonetheless, like in the case of prefix $\langle a,b \rangle$, we obtain a valid firing sequence in the model, that yields us with a marking of the model, i.e. in this case $[p_1] \xrightarrow{\langle t_1, t_2,t_4,t_6 \rangle}[p_6]$.

\paragraph{Generating a Marking-Based Distribution of Possible Next Elements.}
Recall the example alignment depicted in \autoref{fig:_pref_alignment_example}, related to prefix $\langle a,b \rangle$, i.e. the leftmost prefix-alignment.
In marking $[p_3,p_4]$ that corresponds to the prefix-alignment we observe that only transition $t_4$ is enabled, which is labeled $d$.
Therefore, when making a prediction for $\langle a,b \rangle$ on this Petri net we predict next symbol $d$ with probability $1$. However, the task becomes non-trivial when making a prediction for a prefix that corresponds to a marking from which more than one transition in enabled. We now continue by proposing two ways to generate a probability distribution describing the next element for each marking, that deal with the problem of multiple enabled transitions in different ways:
\begin{enumerate}
	\item \emph{Model-driven probability distribution generation}	
	For a marking $m$, we investigate which transitions are enabled in the model, using a uniform distribution over the enabled transitions.
	Based on this uniform distribution over the transitions we calculate the corresponding (not necessarily uniform) categorical distribution over the symbols.	
	\item \emph{Data-driven probability distribution generation}
	For a marking $m$, we determine (not necessarily uniform) categorical distribution over the enabled transitions. Based on this categorical distribution over the transitions we calculate the corresponding categorical distribution over the symbols.
\end{enumerate}

\paragraph{Model-Driven Probability Distribution Generation.}
Consider prefix $\langle a,b,d \rangle$ and the same Petri net, leading to prefix-alignment $\langle (a,t_1), (b,t_2), (d,t_4) \rangle$.
Observe that we fetch corresponding marking  $[p_4, p_5]$.
We observe two enabled transitions from this marking: $t_5$ and $t_6$, Therefore, the uniform distribution over these two transitions describes firing either one of the two with probability $\frac{1}{2}$.
However, $\ell(t_5) = \tau$, i.e. $t_5$ is not observable.
In fact, after firing $t_5$, yielding marking $[p_6]$, we observe that transitions $t_7$, $t_8$ and $t_9$ are enabled, all of which do have an observable label.
Hence, the true set of symbols that can be observed after prefix $\langle a,b,d \rangle$ with non-zero probability is $e$, $f$, $g$ and $h$.
Since we assume the probability distribution over the transitions to be uniform, we observe label $e$ with probability $\frac{1}{2}$, and labels $f$, $g$ and $h$, each with probability $\frac{1}{2} \cdot \frac{1}{3} = \frac{1}{6}$ (i.e., probability $\frac{1}{2}$ to fire $t_5$ from $[p_4,p_5]$ and reach $[p_6]$ and probability $\frac{1}{3}$ for each of the three labels from $[p_6]$).
In the previous example, deriving the exact occurrence probabilities of the different labels is easy, however, in general, it is possible to generate longer transition sequences solely consisting of transitions $t$ with $\ell(t) = \tau$.
In some cases, such sequences can even be of arbitrary length.
We therefore resort to Monte Carlo simulation to approximate the corresponding categorical distribution over the symbols $\Sigma$.

For a given marking $m\in\mathcal{B}(P)$ in an accepting Petri net $\APN$, $\omega(m)=\{t|t\in T \land \bullet t\subseteq m\}$ denotes the set of enabled transitions. 
We, correspondingly, let probability mass function $\mathit{prob}_m:T\rightarrow[0,1]$ assign a firing probability to each transition that is enabled from marking $m$, such that $\Sigma_{t\in\omega(m)}\mathit{prob}_m(t)=1$. 
We assume this probability distribution over the enabled transitions to be a uniform categorical distribution, i.e. $\mathit{prob}^\mathit{uniform}_m(t)=\begin{cases}
\frac{1}{|\omega(m)|}& \text{if } t\in\omega(m),\\\\
0&\text{otherwise}.\\
\end{cases}$
We maintain a counter $c \colon \Sigma \to \mathbb{N}$, with initially $c(a)=0, \forall a{\in}\Sigma$
Starting from marking $m$ in Petri net $\APN$ we pick an enabled transition at random according to probability distribution $\mathit{prob}^\mathit{uniform}_m$. 
Whenever that transition has a corresponding visible label, we count it as the next element, i.e. if $\ell(t)=a$, then $c(a) \gets c(a) + 1$. 
If it relates to an unobservable transition, we fire it, leading to a new marking $m'$ and apply the same procedure, i.e. picking a new enabled transition from  $\mathit{prob}^\mathit{uniform}_{m'}$, up-until we select a transition that has a visible label.
Assume we apply the aforementioned procedure $K$ times (with $K$ the number of Monte Carlo iterations) then the probability of observing  a certain label $a$ is equal to $\frac{c(a)}{K}$.

\paragraph{Data-Driven Probability Distribution Generation.}
We consider the model-driven probability distribution to be the probability distribution that is visually implied by the process model, i.e., the information that the process model visually suggests to the user. However, from an accuracy point-of-view it might be better to fit the categorical distribution over the enabled transitions for each marking, instead of assuming this distribution to be uniform. Therefore, we present the data-driven approach in which we compute an empirical distribution $\mathit{prob}^\mathit{empirical}_m$ after discovering a process model based on the training sequences.
This is rather straightforward: for each prefix of each word in the training log we compute a prefix-alignment to obtain the corresponding marking $m$ in the discovered Petri net $\APN$.
Subsequently, we investigate the transition that we need to fire next, in order to explain the next character observed in the word.
Hence, we base $\mathit{prob}^\mathit{empirical}_m$ on how often each enabled transition $t\in \omega(m)$ was fired when this marking was reached in the training log.\footnote{Alternatively we are able to store a distribution of labels directly in correspondence with marking $m$. However, in such case, the predictor allows us to predict labels which are in fact not described by the process model in the corresponding marking.}
This leads to a probability mass function for each marking in the model that is trained/estimated based on the training data. 
We again apply the same Monte Carlo sampling approach to transform $\mathit{prob}^\mathit{empirical}_m$ into a probability distribution over the next element, instead of over the next transition. 




We have implemented both the Petri-net-based probabilistic classifier based on $\mathit{prob}^\mathit{uniform}_m$ and the one based on the trained $\mathit{prob}^\mathit{empirical}_m$ and they are openly available as part of the ProM process mining toolkit~\cite{Aalst2005} in the package \emph{SequencePredictionWithPetriNets}\footnote{\resizebox{0.975\textwidth}{!}{\url{https://svn.win.tue.nl/repos/prom/Packages/SequencePredictionWithPetriNets/}}}.

Finally, note that, for the purpose of this paper, we assume the fact that given a prefix and a process model, we are able to obtain the corresponding marking in the accepting Petri net.
However, the prediction technique itself is more general and could be used with any alternative approach to obtain a marking in a Petri net given a given a prefix.

\subsubsection{Several Process Discovery Algorithms}
\label{ssec:process-discovery-methods}
Several approaches have been introduced in the process mining field to algorithmically extract a process model from a sequence database (See~\cite{Aalst2016} and~\cite{augusto2018automated} for an overview). 
Here, we introduce the main concepts and ideas behind several process discovery algorithms that we will use in the comparative experiments.

\paragraph{Inductive Miner~\cite{Leemans2013b,Leemans2013}}\mbox{}\\
The Inductive Miner (IM)~\cite{Leemans2013b} is a process discovery algorithm that in a first step extracts a so-called \emph{directly-follows graph} from the sequence database. This directed graph consists of vertices that represent symbols from the log and edges that indicate whether two edges directly follow each other in one of the sequences of the dataset. Edges are annotated with frequency information, i.e., the edge weight corresponds to the number of times one symbol directly follows another symbol. This closely links the directly-follows graph to a 1st-order Markov chain. In a second step, so-called cuts are detected by detecting groups of symbols such that all their connecting edges have the same direction. Based on these cuts, a process model in a tree-based process model notation called a \emph{process tree}~\cite{Buijs2012} is extracted from the directly-follows graph. A process tree is always be transformed to a sound Petri net.

The Inductive Miner infrequent (IMf)~\cite{Leemans2013} is a variant of the IM algorithm that is designed to be able to deal with noisy sequence databases. The IMf algorithm follows the same cut detection procedure as the IM algorithm, but it first filters the directly-follows graph by removing the edges of which the corresponding frequency is less that a certain threshold ratio of the number of sequences, where this threshold is referred to as the \emph{noise threshold}.

\paragraph{Heuristics Miner~\cite{Weijters2011}}\mbox{}\\
The Heuristics Miner (HM)~\cite{Weijters2011} defines a set of heuristics to deduce sequential relations, loop relations, long-term relations, and concurrency relations between symbols in the sequence database. These heuristics are defined in terms of the frequency of certain patterns in the data. The extracted set of sequential, loop, long-term, and concurrency relations are transformed into a process model notation that is called a \emph{heuristics net}. A heuristics net can be transformed into a Petri net, however the resulting Petri net is not guaranteed to be sound: it may contain deadlocks as well as improper completion.

\paragraph{Split Miner~\cite{Augusto2017}}\mbox{}\\
The Split Miner (SM)~\cite{Augusto2017} is similar to the HM algorithm in the sense that it defines a set of heuristics to extract a set of relations between symbols. The name of the algorithm originates from the fact that the extracted relations are used to create a process model in BPMN~\cite{bpmn_2011} notation where the extracted relations are used to determined where in the model the AND-splits (concurrency) and the XOR splits (exclusive choice) should be positioned. The authors show that it is a difficult problem to determine the types of joins corresponding to these splits. Therefore, as a pragmatic solution, the OR-join is used to join all types of splits. This yields a valid BPMN model, but may lead to model with improper completion when the resulting BPMN models are transformed to Petri nets.

\paragraph{ILP Miner~\cite{DBLP:journals/fuin/derWerfDHS09,vanZelst2017_ilp}}\mbox{}\\
In ILP-based process discovery, the sequence database is translated into a prefix-closure, i.e. a set containing all sequence from the database as well as all their prefixes.
The prefix-closure forms the basic set of constraints of an Integer Linear Program (ILP).
The body of constraints makes sure that each solution of such an ILP, corresponds to a place in the resulting Petri net, that allows for all the behaviour observed in the sequence data base. In this way, the process discovery problem is turned into a mathematical optimization problem to find the minimal set of places that are needed to constrain the behavior of the process model to the behavior that was observed in the sequence database.
Several variations of the basic scheme exist, that allow us to pose a variety of formal properties w.r.t. the discovered models.

\paragraph{Evolutionary Tree Miner~\cite{Buijs2012,Buijs2014}}\mbox{}\\
The Evolutionary Tree Miner (ETM)~\cite{Buijs2012,Buijs2014} is a process-tree-based process discovery algorithm, like the IM algorithm. The ETM uses an multi-criteria evolutionary algorithm to optimize towards a process model that scores well on a set of quality criteria. These quality criteria amongst others include \emph{fitness} and \emph{precision} that we introduced in~\autoref{sec:background}.

\paragraph{Indulpet Miner~\cite{Leemans2018_INDULP}}\mbox{}\\
The Indulpet miner is another process-tree-based process discovery algorithm that aims to address the shortcomings of the IM algorithm by combining several existing process tree mining algorithms. The Indulpet miner starts with applying the IM algorithm. For local parts of the process where the IM algorithm fails to deduce any precise process fragment, a novel bottom-up recursion approach is applied, but only for the part of the parts of the dataset that were not already described by the IM. A pattern mining algorithm called the \emph{local process model} (LPM) miner~\cite{Tax2016,Tax2018} to those parts of the dataset that are still not described in a precise manner after the bottom-up procedure. The LPM miner mines frequent patterns of behavior that are expressed in the form of process trees. The ETM algorithm is then used, seeded with the mined set of frequent LPM patterns, to stitch together the patterns into a single model. The process tree notation makes it easy to combine the initial IM model with the models that are created in the later stages: the later models simply become a subtree of the model of the previous stage.

\subsection{Summary of Sequence Modeling Methods and Discussion}
\label{ssec:sequence-model-sumamry}
Many of the sequence models that we discussed in the previous sections conceptually consist of several common sub-procedures: 1) \emph{determining the structure} of the model, thereby creating a set of model states, 2) \emph{optimizing weights} or parameters that determine how these model states map to prediction outputs, and 3) when making prediction for a prefix, map the \emph{prefix to a state} and make the prediction using that state. Different sequence models often differ in the exact algorithms that are used to execute these three sub-procedures. \autoref{tab:relations} gives an overview of the algorithms for these three sub-procedures for several sequence models. 

\begin{table}
	\caption{An overview of several tasks related to sequence models.}
	\label{tab:relations}
	\resizebox{\linewidth}{!}{
		\begin{tabular}{|l|l|l|l|}
			\toprule
			Sequence model & Model structure & \shortstack{State-to-output\\(training time)} & \shortstack{Prefix-to-state\\(prediction time)} \\
			\midrule
			RNN & Hyper-parameter opt.	& SGD + backpropagation & Forward pass\\
			HMM & Hyper-parameter opt. & Baum–Welch algorithm & Viterbi algorithm\\
			Process model & Process discovery & \autoref{sec:petri_net_predictions} & Prefix-alignments\\
			\bottomrule	
	\end{tabular}}
\end{table}

The model structure in the case of RNNs consists of the number of layers, the number of units per layer, and the architecture of the network. These elements are often decided through careful hand-tuning or can alternatively be automated using a hyper-parameter optimization technique. In an HMM this concerns the number of latent variables, which likewise is often selected through hyper-parameter optimization. In the case of a Petri net, the states of the model are its markings. Therefore, \emph{process discovery} can be seen as an automated approach to determining the model structure for a process-model-based predictor. Process discovery can, therefore, be seen as analogous to hyper-parameter optimization in RNNs and HMMs. Like RNNs and HMMs can be hand-tuned instead of automated hyper-parameter optimization, likewise process models can be hand-modeled instead of discovered automatically.


The Baum-Welch algorithm is a well-known algorithm to learn the transition and emission probability parameters of an HMM based on some training data. By determining these parameters, it becomes fixed how a certain state maps to a certain prediction, i.e., according to its emission probabilities. That means that all that is left in order to make a prediction for a given prefix with an HMM is to determine the most likely state for that prefix. For a process-model-based predictor we proposed two approaches in~\autoref{sec:petri_net_predictions} to determine the prediction that should be made for a given model state.

The Viterbi algorithm is a well-known algorithm to determine the most likely sequence of hidden states in an HMM given a prefix or sequence. The last hidden state of such a sequence of hidden states that the Viterbi algorithm returns for a prefix can, therefore, be seen as the most likely state for that prefix. Predictions can be made with an HMM by determining the most likely state for a prefix using Viterbi and then predicting according to the emission probabilities that were provided for that state by the Baum-Welch algorithm during training. In process models, (prefix-)alignments are the common way to determine a sequence of model steps given a prefix or sequence. This creates an analogy between the Viterbi algorithm for HMMs and the alignment algorithm for process models. 

It is important to highlight one crucial difference between Viterbi and alignments: while the Viterbi algorithm provides the \emph{most likely} sequence of hidden states in an HMM given sequence $\sigma$, alignments only provide a sequence of steps in a process model that \emph{deviates the least} from $\sigma$, i.e., no probability information is used to determine the sequence of steps in an alignment. In order to be truly analogous to Viterbi, alignments would need to return not only the sequence of steps through the process model that deviates the least from $\sigma$, but it should additionally need to provide the \emph{most likely} one according to model probabilities when there exist multiple least-deviating sequences of model steps. Several steps have been made towards computing probabilistic alignments~\cite{Alizadeh2014,Alizadeh2015,Koorneef2017}. However, the approaches in~\cite{Alizadeh2014,Alizadeh2015} are heuristic-based and they have been shown to fail to produce the most probable alignment under certain conditions~\cite{Koorneef2017}. The probabilistic alignment approach of~\cite{Koorneef2017} requires not one optimal alignment but instead requires the computation of all optimal alignments, which is computationally intractable. Furthermore, the probabilistic alignment approaches of~\cite{Alizadeh2014,Alizadeh2015,Koorneef2017} have so far not been generalized from alignments to prefix-alignments. Therefore, we resort to non-probabilistic alignments in the experimental comparison in this paper. In the same time, probabilistic prefix-alignments form a relevant research direction in order to develop a true Viterbi-equivalent for process models and to overcome this limitation in the future.

\section{Experimental Setup}
\label{sec:experimental_setup}
\begin{figure}[t]
	\resizebox{\textwidth}{!}{
		\begin{tikzpicture}[
		dir/.tip = {Triangle[scale length=1.5, scale width=1.2]},
		log/.style = {cylinder,shape border rotate=90,aspect=0.25,minimum width=2cm,minimum height=1cm},
		lpm/.style = {minimum width=2cm,minimum height=1cm,double copy shadow={shadow xshift=0.5ex,shadow yshift=-1ex,draw=black!30},fill=white,draw=black},
		model/.style = {minimum width=2cm,minimum height=1cm}
		]
		\node [draw,log] (llog) {Sequence Database};
		\node [draw,log, above right = 0.1 and 1.5 of llog] (train) {Train Sequences};
		\node [draw,log, below right = 0.1 and 1.5 of llog] (test) {Test Sequences};
		\node [draw,model, right= 1.5 of train] (model) {Model};
		\node [inner sep=0pt,outer sep=0pt,minimum size=0,below right = of model] (placeholder) {};
		\node [draw,model, right = 0.5 of placeholder,text width=1.8cm] (predictions) {Predictions for Test\\ Sequences};
		\node [inner sep=0pt,outer sep=0pt,minimum size=0,right = 0.4 of predictions] (placeholder2) {{\normalsize\textbf{(4)}} Evaluate};
		\draw [very thick] 
		(llog.east) edge[-dir] node[left,font=\scriptsize]{} (train)
		(llog.east) edge[-dir] node[above right,font=\scriptsize]{{\normalsize\textbf{(1)}} split} (test)
		(train) edge[-dir] node[below,font=\scriptsize]{{\normalsize\textbf{(2)}} Fit} (model)
		(test.east) edge[bend left=12] (placeholder)
		(model.south) edge[bend left=-23] (placeholder)
		(placeholder.east) edge[-dir] node[below left,font=\scriptsize]{{\normalsize\textbf{(3)}} Apply} (predictions)
		(predictions) edge[-dir] node[right,font=\scriptsize]{} (placeholder2);
		\end{tikzpicture}}
	\caption{An overview of the experimental setup.}
	\label{fig:evaluation_setup}
\end{figure}
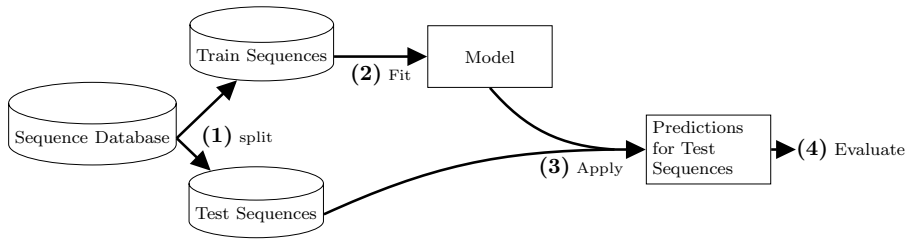
~\autoref{fig:evaluation_setup} gives a high-level overview of the experimental setup that we employ to compare the sequence modeling methods.
First, for each combination of modeling method and sequence database we make a sequence-level random split into $\frac{2}{3}$ training sequences and $\frac{1}{3}$ test sequences. After generating the model on the training sequences we evaluate how well the actual next element predicted for each prefix in the test sequences fits the probability distribution over all possible next elements according to the model. 

For each combination of sequence database and modeling technique  we repeat the experiment three times (with different random splits) to prevent that the results are too dependent on the random sampling of the sequence database into train and test split and over these three results we calculate the 95\% confidence interval around the model performance. 

The performance measure that we use to assess the model performance is called Brier score~\cite{Brier1950}, which is a well-known measure to evaluate a probabilistic classifier and it can intuitively be interpreted as being the mean squared error of the predicted likelihoods over all symbols.

We now continue by giving an overview of the sequence databases used for the evaluation in~\autoref{ssec:sequence-databases} and describing the configurations and the implementations that were used for the sequence models in~\autoref{ssec:methods-configurations}.

\subsection{Sequence Databases}
\label{ssec:sequence-databases}
We evaluate the generalizing capabilities of process discovery techniques and sequence modeling techniques on four real-life sequence databases:
\begin{itemize}
	\item{The \textbf{Receipt phase}\footnote{\url{https://doi.org/10.4121/uuid:a07386a5-7be3-4367-9535-70bc9e77dbe6}} sequence database from the WABO project, containing 8577 events of 27 symbols originating from 1434 cases of the receipt phase of the building permit application process at a Dutch municipality.}
	\item{The \textbf{BPI'12}\footnote{\url{https://doi.org/10.4121/uuid:3926db30-f712-4394-aebc-75976070e91f}} sequence database which contains cases from a financial loan application process at a large financial institute, consisting of 164506 events divided over 13087 sequences and 23 symbols}
	\item{The \textbf{SEPSIS}\footnote{\url{https://doi.org/10.4121/uuid:915d2bfb-7e84-49ad-a286-dc35f063a460}} sequence database~\cite{Mannhardt2017}, containing medical care pathways of 1050 sepsis patients, for which in total 15214 events were logged from 16 different symbols.}
	\item{The \textbf{NASA}\footnote{\url{https://doi.org/10.4121/uuid:60383406-ffcd-441f-aa5e-4ec763426b76}} sequence database~\cite{Leemans2018}, which contains method-call-level events that each describe a single run of an exhaustive unit test suite for the NASA Crew Exploration Vehicle (CEV)\footnote{\url{http://babelfish.arc.nasa.gov/hg/jpf/jpf-statechart}}. The dataset consists of 2566 sequences consisting of 36819 events in total over 47 symbols.}
\end{itemize}

\subsection{Configurations and Implementations}
\label{ssec:methods-configurations}
We apply all the sequence modeling techniques that we introduced in \autoref{sec:prediction}. Additionally, we include a sequence compression algorithm in the evaluation. It has been shown~\cite{Weinberger1997} in the information theory field that the tasks of prediction and compression are closely related, and that good compression methods are also good prediction methods. Gopalratnam and Cook~\cite{Gopalratnam2007} adapted the well-known LZ78 compression algorithm~\cite{Ziv1978} to make it usable in a predictive way, calling the prediction approach Active LeZi.

Most of the sequence modeling methods have several hyper-parameters that can be manually selected or automatically tuned in order to achieve good performance on a given dataset. For instance, in the case of HMM, the number of hidden states needs to be selected. The performance of the sequence modeling methods (AKOM, HMM, RNN, GRU, and LSTM) can highly depend on the chosen configuration of hyper-parameters. Therefore, we conduct an optimization procedure for these methods to find the best-performing parameter setting before building the final model. To this end, we split the set of training sequences further into two parts using a random split, resulting in a 80\% inner training set (53\% of original sequences) and a 20\% validation set (13\% of original sequences). We then test multiple parameter configurations by training the model on the inner training set and measuring the performance on the validation set. We choose the configuration that achieved the best Brier score on the validation set and use these parameter settings to build the final model on all training sequences.

The hyper-parameters included in the optimization procedure and the considered values for these parameters are shown in~\autoref{tab:hyperparameters}. As the parameter space for AKOM, HMM, and the abstraction-based approach is rather small, we perform a grid search for these methods, testing all the possible combinations of the considered values. In case of the neural-network-based models we employ a state-of-the-art hyper-parameter optimization that is called tree-structured Parzen estimator (TPE)~\cite{bergstra2011algorithms}. The TPE optimizer is necessary for the case of neural networks since the high dimensionality its hyper-parameter search space makes a grid search impractical. TPE is a sequential model-based optimization procedure that in each iteration selects a parameter configuration for testing according to predefined distributions for each parameter. 

\begin{table}[tb]
	\centering
	\caption{Definition of the hyper-parameter search spaces for the sequence modeling methods.}
	\label{tab:hyperparameters}
	\resizebox{\linewidth}{!}{
		\begin{tabular}{|l|l|l|r|}
			\toprule
			Method & Procedure & Parameter & Considered values \\
			\midrule
			\multirow{6}{*}{RNN, GRU, LSTM} & \multirow{6}{*}{TPE} & $\mathit{n\_layers}$ & $\{1, 2, 3\}$ \\
			&  & $\mathit{layer\_size}$ & $[10, 150]$ \\
			&  & $\mathit{batch\_size}$ & $\{2^3, 2^4, 2^5, 2^6\}$ \\
			&  & $\mathit{dropout}$ & $[0, 0.5]$ \\
			&  & $\mathit{l_1}$ & $[0.00001, 0.1]$ \\
			&  & $\mathit{l_2}$ & $[0.00001, 0.1]$ \\
			\midrule
			AKOM & grid search & $k$ & $\{1, 2, ..., 19\}$ \\
			\midrule
			\multirow{2}{*}{HMM}  & \multirow{2}{*}{grid search} & $n\_\mathit{states}$ to $|\Sigma|$ ratio & $\{0.1, 0.5, 1.0, 1.5, 2.0, 3.0, 5.0\}$ \\
			&  & $\mathit{regularizer}$ & $\{\mathit{None}, l_2, l_\infty\}$ \\
			\midrule
			\multirow{2}{*}{Automaton-based} & \multirow{2}{*}{grid search} & $\mathit{type}$ & $\{\mathit{seq,mult,set}\}$ \\ &&$k$ & $\{1, 2, ..., 19\}$ \\
			\bottomrule
	\end{tabular}}
\end{table}
\subsubsection{Neural Networks}
For the neural-network-based approaches (RNN, GRU, and LSTM) we use their respective implementations provided in the Keras\footnote{\url{https://keras.io}} Python deep learning library. For each of these types of neural networks we explore multiple architectures described later in this subsection. We optimize the weights of the models using Adam~\cite{Kingma2015}, which is a recent variant of stochastic gradient descent that has empirically shown te perform well. Furthermore, we apply early stopping, i.e., we stop training the neural network if no performance improvement has been seen for 20 training iterations in a row. Early stopping can help to reduce or prevent overfitting by preventing the model weights to take values that represent specific characteristics of the training data that do not generalize to the test data. Architecture choices such as the number of layers of the neural network and the number of units per layer are left to the hyper-parameter optimization procedure (as shown in~\autoref{tab:hyperparameters}, together with the learning rate parameter which determines how large steps the gradient descent weight optimization procedure takes in the direction of the gradient. Finally, the hyper-parameter optimization procedure includes dropout, $l_1$ and $l_2$, which are three types of model regularization that can help to prevent overfitting by punishing overly complex weight structures of the neural network, thereby applying Occam's razor and steering the neural network towards simpler solutions. The code to train and evaluate the neural network architectures is available in a Github repository that accompanies this paper.\footnote{\label{ft:github}\url{https://github.com/TaXxER/rnnalpha}}

\subsubsection{Markov Models}
For the Markov chains we apply the first-order as well as second-order Markov chain. 
We have implemented the Markov chain predictors and the AKOM model ourselves and the code is publicly available in the Github repository.\textsuperscript{\ref{ft:github}} For the AKOM model we optimize the hyper-parameter $k$ using grid search.
For Hidden Markov Models (HMM), we use the \texttt{hmm.discnp}\footnote{\url{https://CRAN.R-project.org/package=hmm.discnp}} library in R. We perform a grid search hyper-parameter optimization to select the number of hidden states of the HMM as well as the type of regularization that is used to prevent the model from overfitting.

\subsubsection{Grammar Inference}
For the spectral learning grammar inference technique we use the implementation that is provided in the Python spectral learning toolkit Scikit-SpLearn~\cite{Arrivault2017}. The implementation of train-test procedure that makes use of this code is included in the Github repository that accompanies this paper.~\textsuperscript{\ref{ft:github}}

\subsubsection{Automaton-based Approaches}
The automaton-based next-element prediction techniques have been implemented in the Python process mining library pm4py\footnote{\url{http://pm4py.org/}}. We perform a grid search hyper-parameter optimization to select the type of abstraction that is used for the automaton (i.e., set, multiset, or sequence abstraction) and the window size parameter $k$.

\subsubsection{Process Model Approaches}
For the next-element predictors based on Petri nets we apply the process discovery techniques that were described in~\autoref{ssec:process-discovery-methods} with their default parameter settings, unless were we will explicitly state a different parameter value. Hyper-parameter optimization of the process discovery techniques for process-model-based prediction is not possible due to the computational time needed to make process-model-based prediction (mainly due to the computation time needed to calculate prefix-alignments). However, in general, hyper-parameters are not as omnipresent for process discovery approaches as opposed to machine learning techniques. For hyper-parameters that are of vital importance to the success of process discovery, such as the frequency threshold of the Inductive Miner, we will test and report the performance using several settings. The experiments with the process-model-based predictors are performed using their implementation in \emph{SequencePredictionWithPetriNets}\footnote{\resizebox{0.975\textwidth}{!}{\url{https://svn.win.tue.nl/repos/prom/Packages/SequencePredictionWithPetriNets/}}} package of the ProM process mining toolkit~\cite{Aalst2005}.

\section{Results}
\label{sec:results}

\begin{table}[tb]
	\centering
	\caption{The mean of Brier score and the 95\% confidence interval (ranging from $\mu\pm\mathit{CI}$) for each combination of method and dataset.}
	\label{tab:results}
	\resizebox{\linewidth}{!}{
		\begin{tabular}{|l|w|W|b|B|s|S|n|N|}
			\toprule
			Method & \multicolumn{2}{|c|}{Receipt Phase} & \multicolumn{2}{|c|}{BPI'12}& \multicolumn{2}{|c|}{SEPSIS}& \multicolumn{2}{|c|}{NASA}\\
			\midrule
			& \multicolumn{1}{c|}{$\mu$} & \multicolumn{1}{c|}{CI} & \multicolumn{1}{c|}{$\mu$}& \multicolumn{1}{c|}{CI} & \multicolumn{1}{c|}{$\mu$} & \multicolumn{1}{c|}{CI}& \multicolumn{1}{c|}{$\mu$} & \multicolumn{1}{c|}{CI}\\
			\midrule
			\multicolumn{9}{|c|}{\emph{Baselines Methods}}\\
			\midrule
			Random guessing 					& 0.0381 & 0.0010&0.0417 & 0.0000&0.0620 & 0.0061 &0.0213 & 0.0001\\
			Proportional guessing 				& 0.0336 & 0.0009&0.0402 & 0.0000&0.0542 & 0.0002 &0.0209 & 0.0001\\
			\midrule
			\multicolumn{9}{|c|}{\emph{Process Mining: Process Discovery with Uniform Distribution per Marking}}\\
			\midrule
			IM~\cite{Leemans2013b} 				& 0.0338 & 0.0018&0.0314 & 0.0013&0.0612 & 0.0053 &0.0207 & 0.0003\\
			IMf 20\%~\cite{Leemans2013} 		& 0.0224 & 0.0015&0.0293 & 0.0001&0.0486 & 0.0011 &0.0152 & 0.0001\\
			IMf 50\%~\cite{Leemans2013} 		& 0.0191 & 0.0037&0.0390 & 0.0005&0.0759 & 0.0085 &0.0211 & 0.0006\\
			HM~\cite{Weijters2011}				& 0.0245 & 0.0008&0.0258 & 0.0004&0.0442 & 0.0005 &0.0177 & 0.0004\\
			SM~\cite{Augusto2017}				& 0.0262 & 0.0022&0.0252 & 0.0002&0.0574 & 0.0026 &0.0160 & 0.0002\\
			ILP~\cite{vanZelst2017_ilp}	& 0.0167 & 0.0012&0.0413 & 0.0037&0.0526 & 0.0013 &0.0232 & 0.0013\\
			ETMd~\cite{Buijs2012} 				& 0.0196 & 0.0028&0.0287&0.0028&0.0526&0.0013&0.0197 & 0.0004\\
			Indulpet~\cite{Leemans2018_INDULP}			& 0.0249 & 0.0010&0.0429 & 0.0059&0.0578 & 0.0024&0.0205 & 0.0043\\
			\midrule
			\multicolumn{9}{|c|}{\emph{Process Mining: Process Discovery with Trained Distribution per Marking}}\\
			\midrule
			IM~\cite{Leemans2013b} 				& 0.0255 & 0.0027&0.0287 & 0.0011&0.0455 & 0.0035 &0.0202 & 0.0002\\
			IMf 20\%~\cite{Leemans2013} 		& 0.0152 & 0.0015&0.0293 & 0.0001&0.0395 & 0.0014 &0.0106 & 0.0003\\
			IMf 50\%~\cite{Leemans2013} 		& 0.0153 & 0.0009&0.0347 & 0.0009&0.0664 & 0.0017 &0.0207 & 0.0005\\
			HM~\cite{Weijters2011}				& 0.0181 & 0.0007&0.0231 & 0.0003&0.0372 & 0.0013 &0.0159 & 0.0006\\
			SM~\cite{Augusto2017}				& 0.0099 & 0.0006&0.0226 & 0.0001&0.0513 & 0.0026 &0.0155 & 0.0002\\
			ILP~\cite{vanZelst2017_ilp} 	& 0.0167 & 0.0012&0.0445 & 0.0059&0.0512 & 0.0013 &0.0232 & 0.0012\\
			ETMd~\cite{Buijs2012} 				& 0.0114 & 0.0010&0.0263&0.0059&0.0396&0.0013 &0.0195 & 0.0002\\
			Indulpet~\cite{Leemans2018_INDULP}			& 0.0153 & 0.0041&0.0441 & 0.0050&0.0451 & 0.0102&0.0195 & 0.0025\\
			\midrule
			\multicolumn{9}{|c|}{\emph{Process Mining: Automata Based prediction}}\\
			\midrule
			Automaton-based (\autoref{ssec:methods:automaton})									&0.0072 & 0.0002&0.0120 & 0.0000&0.0283 & 0.0004&0.0052 & 0.0000\\
			\midrule
			\multicolumn{9}{|c|}{\emph{Machine Learning: Neural Networks}}\\
			\midrule
			RNN									& 0.0072 & 0.0007&0.0159 & 0.0003&0.0277 & 0.0000 &0.0048 & 0.0001\\
			LSTM								& 0.0075 & 0.0012&0.0122 & 0.0000 &0.0277 & 0.0008&0.0049 & 0.0002\\
			GRU									& 0.0073 & 0.0008&0.0127 & 0.0001&0.0277 & 0.0004&0.0048 & 0.0000\\
			\midrule
			\multicolumn{9}{|c|}{\emph{Machine Learning: Compression}}\\
			\midrule
			Active LeZi~\cite{Gopalratnam2007} 	&0.0128 & 0.0004&0.0182 & 0.0011&0.0331 & 0.0002&0.0088 & 0.0004\\
			\midrule
			\multicolumn{9}{|c|}{\emph{Markov Models}}\\
			\midrule
			1st-order Markov chain 				& 0.0114 & 0.0007 & 0.0207 & 0.0000 & 0.0342 & 0.0002&0.0066 & 0.0000\\
			2nd-order Markov chain 				& 0.0110 & 0.0001 & 0.0135 & 0.0000& 0.0313 & 0.0003 &0.0052 & 0.0000\\
			AKOM~\cite{Pitkow1999} 				& 0.0070 & 0.0000 & 0.0119 & 0.0000& 0.0262 & 0.0003 &0.0049 & 0.0000\\
			Hidden Markov Model 				& 0.0184 & 0.0019 &	0.0188 & 0.0017& 0.0340 & 0.0009 &0.0081 & 0.0009\\
			\midrule
			\multicolumn{9}{|c|}{\emph{Grammar Inference}}\\
			\midrule
			Spectral Learning~\cite{Balle2014}	&0.0195 & 0.0001 &0.0370 & 0.0004&0.0480 & 0.0010 &0.0207 & 0.0002\\
			\bottomrule
	\end{tabular}}
\end{table}
\autoref{tab:results} shows the results in terms of Brier score on the four datasets for each of the techniques. The worst Brier score value of each $\mu$-column in the table is colored in red and the best value is colored green, with the other values taking an intermediate color. In the CI columns, the color represent the consistency of the approach: if the 95\%-CI range has a small width (i.e., if the method performed very consistently amongst the three runs), then the cell is colored in green and otherwise in red. Two baseline methods are included in the table: the \emph{random guessing} baseline corresponds to predicting the equal probability to each symbol (i.e., predicting according to a uniform categorical distribution), while the \emph{proportional guessing} baseline corresponds to predicting according to the frequency distribution of symbols in the training sequences. 

Overall, the three neural network types, AKOM, and the automaton-based approach have the lowest error in terms of Brier score, with only very small differences between their accuracies. AKOM is the best performing sequence model on average on three of the four datasets, with GRU being the best performing sequence model on average on the NASA software log. The confidence interval around the mean Brier score for the neural network methods turns out to be wider than for AKOM and the automaton-based predictor, meaning that while their mean Brier scores are similar, the neural networks were impacted to a larger degree by the random splits into training and test data. This might indicate that the neural network approaches are more prone to overfitting the training data, even though we applied regularization to prevent overfitting and used hyper-parameter optimization to select the degree of regularization.

On all four datasets the Brier scores for the aforementioned methods are considerably better than for the process-model-based approaches and than grammar inference, meaning that they provide considerably more accurate probability distributions over the next event for prefixes from previously unseen sequences. This finding is independent of whether uniform categorical distributions or trained categorical distributions were used for the process models. 


The results also show that learning a categorical probability distribution over the enabled transitions for each marking from the training data leads to more accurate predictions on the test data compared to the approach where we assumed the categorical probability distribution over the enabled transitions to be uniform. Note, however, that in the process mining field the discovered process models are often used to communicate with process stakeholder about the business process, and that the process model discovered with process discovery typically have no branching probabilities shown in the model. Therefore, one could say that the uniform distribution matches the graphical representation of the Petri net. The Split Miner~\cite{Augusto2017} is the best performing process discovery technique on two of the four logs when we learn the probability distribution per marking from the training data, with on the other logs the Heuristics Miner~\cite{Weijters2011} and the Inductive Miner with 20\% filtering~\cite{Leemans2013} being the best approach. An interesting observation can be made about the Indulpet Miner~\cite{Leemans2018_INDULP}, which does not perform well on average performs but has a very large 95\%-CI for all three logs, indicating that for some of the random train/test-splits the method generates quite accurate predictions but for others very inaccurate ones.

The spectral learning grammar inference performs similarly to the process discovery techniques when training the distribution per marking. This shows that the techniques from the grammar inference field and from the process mining field, which both have the aim to generate human-interpretable sequence models, are less accurate than the machine learning techniques that have no goal of interpret-ability and focus solely on accuracy.

The prediction accuracy of the automaton-based techniques from the process mining field is remarkable, since these techniques are still somewhat interpretable: the simple nature of the abstraction functions that we introduced in~\autoref{ssec:methods:automaton} imply that the resulting automaton has interpretable states. Furthermore, techniques exists to transform this automaton into a Petri net (see~\cite{DBLP:journals/sosym/AalstRVDKG10}). Note that in some sense, AKOM could also be argued to be an interpretable model in the sense that for a given prediction it can be easily be traced back what caused the model to make this prediction by looking up the state in the Markov model that was used to make the prediction. However, if we put the threshold for interpretability at a stronger notion of interpretability: "can we get insight into the model behavior by looking at the model" instead of "can we trace back the reason why a model made a certain prediction", then AKOM does not yield an interpretable model since it consists of $k$ individual Markov models that would each need to be comprehended to understand the understand the model behavior. The automaton-based predictors based on set, multiset, and sequence abstraction fit both notions of interpretability.

\section{Related Work}
\label{sec:related_work}
We group related work into several directions of related work. Measures for generalization from the process mining field are one area of related work, which we discuss in~\autoref{ssec:generalization}. Another area of related work is predictive business process monitoring, which we discuss in~\autoref{ssec:pred-bus}. Finally, in~\autoref{ssec:non-probabilistic} we discuss several sequence prediction techniques that predict only the single most likely next element instead of predicting a probability distribution over all possible next elements.


\subsection{Measuring Generalization}
\label{ssec:generalization}
The work presented in this paper is closely related to the challenge of measuring precision in process mining. In the process mining field, generalization is often defined as ``the likelihood that the process model is able to describe yet unseen behavior of the observed system''~\cite{Buijs2012}. This definition is noticeably different from the definition of generalization in the machine learning field. The process mining definition of generalization is an asymmetric one: it specifies the model should ideally allow for sequences from the test set, but it does not specify that the models should \emph{not} allow for sequences that are \emph{not} in the test set. A consequence of this definition is that a model that allows for all behavior is the most generalizing one. In contrast, in the machine learning field it is common to have a more probabilistic notion of generalization: the model should specify a probability distribution over sequences that make sequences in the test set likely (and as a probability distribution has to sum to 1, an effect is that other sequences should be unlikely). 

Several generalization measures have been proposed in the process mining field that quantify the degree to which a given process model generalizes the behavior that is observed in a given sequence dataset~\cite{Aalst2012,Broucke2014,Dongen2016}. All of these measures calculate the generalization of the process model \emph{with respect to the same sequence database from which the process model was discovered}. In contrast, in the machine learning field it is common to measure generalization by splitting the data into a separate \emph{training} set that is used to learn the model and a \emph{test} set on which it is evaluated how well the model fits this data. Because the test set is disjoint from the training set, the fit between model and test set can be considered to measure the \emph{generalization} of the model to the test data.

\subsection{Predictive Business Process Monitoring}
\label{ssec:pred-bus}
The predictive business process monitoring is a research field that is concerned with sequence predictions within the application domain of \emph{business process management}. The field focuses on several prediction tasks for ongoing instances of a business process, including prediction of an outcome of the process instance~\cite{di2016clustering,Teinemaa2016}, prediction of the remaining time of a process instance~\cite{Dongen2008,Rogge2015,Tax2017}, predicting deadline violations~\cite{pika2012predicting}, and, most relevant to this work, prediction of the coming business activities with a running instance of a business process~\cite{van2012process,Tax2017}.

Several approaches have been proposed in the literature to tackle the challenge of predicting the next business activity in ongoing instances of a business process. Often events are considered to be multi-dimensional, i.e., there are additional attributes that describe the events in addition to only the business activity. Some of the methods encode the available data as a feature vector, and subsequently apply a classifier to predict the next event~\cite{pravilovic2013process,unuvar2016leveraging,marquez2017run}.
Other techniques discover a process/sequence model from the control-flow using sequential pattern mining~\cite{ceci2014completion}, Markov models~\cite{lakshmanan2015markov} or a Probabilistic Finite Automaton~\cite{breuker2016comprehensible}. 
Often, as a second step after discovering the process model, classifiers are built for each state in the model, enabling the inclusion the data payload of the ongoing case into the prediction process~\cite{ceci2014completion,lakshmanan2015markov}. In~\cite{DBLP:conf/bpm/SchonenbergWDA08}, the authors propose a recommendation engine that allows to predict the business activity based on the assumption that a process model of the underlying process is known. As a consequence, a collection of possible next elements is assumed to be known/given.
The recommendation engine allows the user to compute the best possible next element from the given collection, that is expected to most positively impact a user-specified KPI.

Rogge-Solti developed several techniques~\cite{Rogge2013a,Rogge2013b,Rogge2015} to predict the remaining cycle time using stochastic Petri nets (SPN) and generalized stochastic Petri nets (GSPN)~\cite{Ajmone1984}. These SPNs and GSPNs closely link to the Petri nets with probability that we introduce in this work, however, there is an important difference: where the Petri nets in our work define a categorical probability distribution over the next transition, SPNs and GSPNs define a continuous probability distribution that specifies the \emph{timing} of transition firings. SPNs and GSPNs thereby only implicitly specify an ordering: transitions that are likely to fire soon are more likely to be the next transition to fire. Furthermore, the work of Rogge-Solti applies these SPN and GSPN models only for the prediction of the remaining cycle time and does not address next-element prediction for unfinished sequences.

More recently, deep learning approaches, i.e. in particular Long Short-Term Memory (LSTM) unit based networks, have been applied in the context of next element prediction~\cite{Evermann2017,Mehdiyev2017,Tax2017}. 
However, none of these studies compare their method to existing process discovery techniques. 
Therefore, in this work we aim to bridge this gap by comparing the most widely used representatives from the sequence modeling field to well-known process discovery techniques. 
Furthermore, while most of the next element prediction methods strive for a high accuracy for a given process instance, our focus in this paper is on assessing the generalizing capabilities of the sequence modeling/process discovery techniques in terms of control-flow.

\subsection{Non-probabilistic Sequence Classification}
\label{ssec:non-probabilistic}
In this paper we have focused on sequence models where the next-element predictions are probabilistic, i.e., that provide the probability distribution over the set of possible outcomes instead of simply giving the single most likely outcome. Several methods from the data mining community focus solely on predicting the single most likely next element of a sequence without generating the whole probability distribution, i.e., non-probabilistic sequence models. 

One of such non-probabilistic sequence prediction algorithms is the \emph{compact prediction tree} (CPT)~\cite{Gueniche2013} algorithm, which was later improved to the computationally more efficient CPT+ algorithm~\cite{Gueniche2015}. The CPT and CPT+ algorithms generate a tree-based data structure that makes a lossless compression from the training data which can be used at prediction time to efficiently search for the most likely next element.

\section{Conclusions \& Future Work}
\label{sec:conclusion}

In this work we proposed two techniques to use process mining techniques as generative sequence models we have introduced two ways in which (discovered) Petri nets can be used as probabilistic classifiers to predict the next element for a given prefix of a sequence: a uniform distribution approach, which uses only the information that is visually communicated by the graphical representation of the Petri net, and an empirical distribution approach that optimizes a categorical probability distribution per marking using a training log. 

We have used these two techniques to use process mining methods as generative sequence models to perform a comparison of sequence modeling techniques on a collection of four real-life sequence databases that spans three research fields: \emph{grammar inference}, \emph{process mining} and \emph{machine learning}. To the best of our knowledge, there has so far been no comparative evaluation that compares sequence modeling techniques from these different fields. Techniques from the grammar inference field and from the process mining field have an aim to generate human-interpretable sequence models, where machine learning methods often have no aim to be interpretable and focus solely on accurate predictions. We have found that overall, the black-box techniques from the machine learning field generate more accurate predictions than the interpretable models from the grammar inference and from the process mining fields. This shows that machine learning sequence modeling might be a better choice than process discovery methods to model a business process when interpretability of the model is not a requirement.

The main limitation of the current approach of process-model-based predictions lies in the non-probabilistic nature of prefix-alignments. We see probabilistic prefix-alignments as a vital future area of research in order to more accurately infer the categorical probability distributions over the next symbols for each marking of the process model.

\bibliographystyle{spbasic}
\bibliography{bibitems}

\end{document}